\documentclass[journal]{IEEEtran}
\usepackage{graphics} 
\usepackage{graphicx}
\usepackage{epsfig} 
\usepackage{times} 
\usepackage{amsmath} 
\usepackage{amssymb}  
\usepackage{amsfonts}
\usepackage{xcolor}
\usepackage{siunitx}
\usepackage{amsthm}
\usepackage{hyperref}
\usepackage{tabularx}
\usepackage{balance}
\usepackage{placeins}
\usepackage{layout}
\usepackage{pifont}

\usepackage{soul} 

\usepackage{enumitem}

\usepackage{mymacros}

\newtheorem{theorem}{Theorem}

\newcommand{\cmark}{\ding{51}}%
\newcommand{\xmark}{\ding{55}}%

\begin{document}

\title{Safe Control with Learned Certificates: A Survey of Neural Lyapunov, Barrier, and Contraction Methods for Robotics and Control}

\author{Charles~Dawson, Sicun~Gao, 
        and~Chuchu~Fan
\thanks{C. Dawson and C. Fan are with the Department
of Aeronautics and Astronautics at the Massachusetts Institute of Technology.
This work was supported by the Defense Science and Technology Agency in Singapore, but this article solely reflects the opinions and conclusions of its authors and not DSTA Singapore or the Singapore Government. C. Dawson is supported by the NSF GRFP under Grant No. 1745302.}
\thanks{S. Gao is with the Department of Computer Science and Engineering at the University of California, San Diego.}}

\maketitle

\begin{abstract}
Learning-enabled control systems have demonstrated impressive empirical performance on challenging control problems in robotics, but this performance comes at the cost of reduced transparency and lack of guarantees on the safety or stability of the learned controllers. In recent years, new techniques have emerged to provide these guarantees by learning certificates alongside control policies --- these certificates provide concise, data-driven proofs that guarantee the safety and stability of the learned control system. These methods not only allow the user to verify the safety of a learned controller but also provide supervision during training, allowing safety and stability requirements to influence the training process itself. In this paper, we provide a comprehensive survey of this rapidly developing field of certificate learning. We hope that this paper will serve as an accessible introduction to the theory and practice of certificate learning, both to those who wish to apply these tools to practical robotics problems and to those who wish to dive more deeply into the theory of learning for control.
\end{abstract}


\section{Introduction}

\begin{figure*}[t]
\centering
\includegraphics[width=0.9\linewidth]{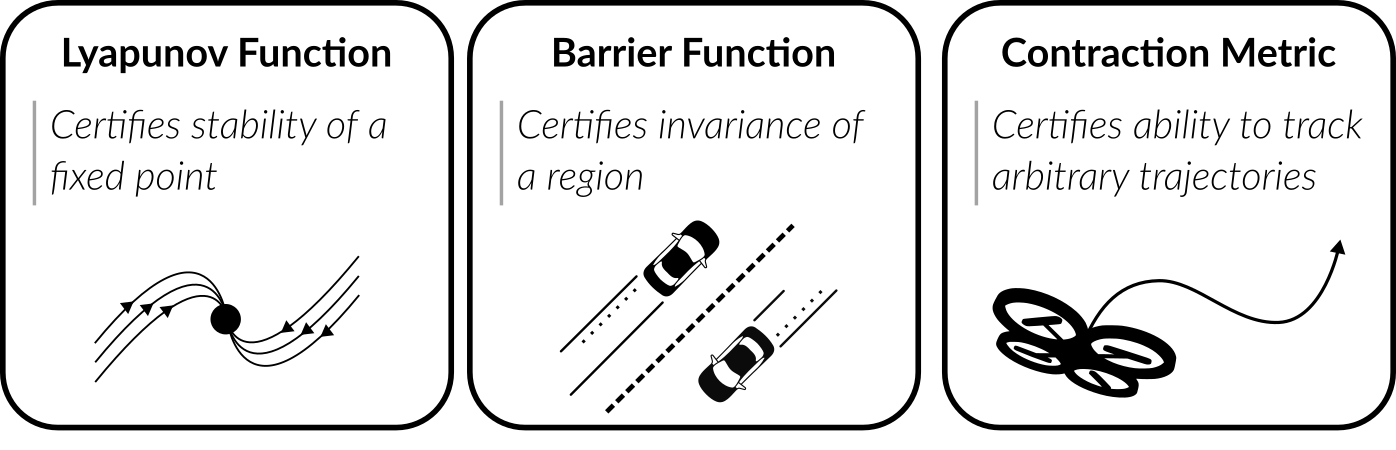}
\caption{Lyapunov functions, barrier functions, and contraction metrics are three common types of control certificate. Each can be used to certify different properties.}
\label{fig:headline}
\end{figure*}

\IEEEPARstart{M}{any} desirable properties of dynamical systems, including stability, safety, and robustness to disturbance, can be proven via the use of \textit{certificate functions}. Perhaps the most well known of these certificate functions are Lyapunov functions: pseudo-energy functions that, if shown to be strictly decreasing along trajectories of the system, prove the stability of the system about a fixed point \cite{Giesl2015}. Other well-known certificates include barrier functions (which prove forward invariance of a set, and thus safety \cite{Taylor2019,Peruffo2020}) and contraction metrics (which prove differential stability in trajectory tracking \cite{Singh2020,tsukamoto21survey}). A summary of these different certificates is shown in Fig.~\ref{fig:headline}.

These certificate functions can be extremely valuable to the control system designer, as they allow her to prove safety and stability properties even for complex and nonlinear control systems. However, although the theory of certificates like Lyapunov functions is more than a century old, it is not until the last decade that general numerical methods have emerged to construct certificates, and even then many proposed methods were computationally intractable (e.g. relying on solving a high-dimensional PDE numerically)~\cite{Giesl2015}. Absent efficient general-purpose methods, finding certificates requires spending great effort to hand-design certificates for specific systems. Even in the best case, this hand-tuning requires a good deal of intuition and luck to find an appropriate functional form (e.g. fixed-degree polynomial) and parameters (e.g. polynomial coefficients) for the certificate.

In recent years, new techniques have emerged for automatically synthesizing certificate functions. For systems with polynomial dynamics, the search for a valid certificate can be framed as a convex semi-definite optimization problem through the use of sum-of-squares (SoS) techniques \cite{Ahmadi2016}. Unfortunately, SoS methods are limited to polynomial systems and scale poorly to higher dimensional systems~\cite{Srinivasan2021ExtentCompatibleCB}. To address these shortcomings, an emerging body of work in the control theory, machine learning, and robotics communities have employed neural networks to learn approximate certificate functions \cite{Chang2019}.

Unlike traditional approaches to learning for control that search only for a control policy (such as many reinforcement learning, or RL, methods), certificate-based techniques simultaneously search for a control policy and a certificate that proves the soundness of that policy. This search may be guided by a separate reward function, similarly to traditional RL methods~\cite{Chang2021}, but the reward function can be omitted entirely in favor of a self-supervised training signal provided by the certificates themselves~\cite{Chang2019,dawson2021safe,Qin2021,xiao2021barriernet}. Because they produce a verifiable correctness certificate alongside a learned control policy, these \textit{neural certificates} provide a trustable approach to learning for control, helping to address concerns about the safety and reliability of learned controllers. Neural certificates have been successfully applied to complex nonlinear control tasks, including stable walking under parametric uncertainty \cite{Choi2020}, precision quadrotor flight through turbulence induced by propeller wash \cite{Sun2020}, and provably safe decentralized control of multi-agent systems containing over 1,000 agents \cite{Qin2021}.

\subsection{Scope and Contributions}

This is a rapidly developing field for which no comprehensive survey has yet been written. Giesl and Hafstein's 2015 review covers traditional methods for Lyapunov function synthesis, focusing on techniques for finding Lyapunov functions as the solutions of partial differential equations or through linear and second-order cone programming, but does not cover neural representations~\cite{Giesl2015}. Ahmadi and Majumdar's 2016 paper presents a relevant summary of the sum-of-squares optimization-based approaches for synthesizing Lyapunov functions, but similarly does not cover neural representations (and only applies to systems with polynomial dynamics~\cite{Ahmadi2016}). The 2021 review of contraction theory by Tsukamoto, Chung, and Slotine~\cite{tsukamoto21survey} discusses neural approaches to learning one particular type of control certificate, but does not discuss the most common certificates (Lyapunov and barrier functions). Our goal is to provide a comprehensive survey of recent developments in certificate learning to serve as an accessible introduction to these tools, both to those who might wish to apply them to practical robotics problems and to those who wish to dive more deeply into the theory of learning for control. Our survey includes theoretical discussion of all three common types of control certificate: Lyapunov functions, barrier functions, and contraction metrics. However, reflecting the prevalence of Lyapunov and barrier functions in the literature, our case studies focus mainly on these two types of certificate (readers interested in case studies involving contraction metrics are referred to the tutorial paper by Tsukamoto, Chung, and Slotine~\cite{tsukamoto21survey}). Our review is organized as follows.
\begin{itemize}
    \item Section~\ref{background} provides the relevant background from control theory, introducing Lyapunov functions, barrier functions, and contraction metric certificates.
    \item Section~\ref{related-work} discusses prior, non-neural approaches to certificate synthesis.
    \item Section~\ref{learning} discusses how neural networks can be used to synthesize certificates and their corresponding safe controllers. This section attempts to unify the various synthesis methods presented in the literature, but also includes some historical discussion.
    \item Section~\ref{implementation} discusses issues that arise when implementing certificate-based controllers on hardware, and Section~\ref{case-studies} presents a number of case studies applying these techniques to practical robotic systems, both in simulation and hardware.
    \item Finally, Sections~\ref{future_work} and~\ref{conclusion} concludes by discussing some open problems in this area.
\end{itemize}

\section{Background}\label{background}

This section provides the necessary background from control theory by outlining the various types of certificate function and their use in controller design. Readers with a background in control theory should feel free to skim this section. We begin by defining notation and terminology, then examine the three major types of control certificates: Lyapunov functions, barrier functions, and contraction metrics.

\subsection{Dynamical systems}

In the rest of this paper, we will consider general dynamical systems of the form $\dot{x} = f(x, u)$, where $x \in \X \subseteq \R^n$ is the state, $u \in \U \subseteq \R^m$ is the input, and $f: \X \times \U \mapsto \X$ is the flow map (which we assume to be locally Lipschitz in $x$ and $u$). The sets $\X$ and $\U$ represent the sets of admissible state and control inputs, respectively. Often, we will consider a restricted (but still quite general) class of dynamics known as \textit{control-affine}: $\dot{x} = f(x) + g(x)u$, where $g: \R^n \mapsto \R^{n\times m}$ is also assumed to be locally Lipschitz. For clarity, we will focus only on the continuous time case, but the theory for the discrete-time case follows similarly; readers interested in the discrete-time theory may read~\cite{Agrawal2017DiscreteCB} or~\cite{tsukamoto21survey}.

In the context of these dynamics, the controls engineer's task is to find a feedback controller $\pi: \X \mapsto \U$ such that the control input $u = \pi(x)$ imparts the \textit{closed-loop system} $\dot{x} = f_c(x) = f(x, \pi(x))$ with certain desirable properties (e.g. stability). In general, $\pi$ may be a function of time as well as state, as in trajectory-tracking control. For clarity, we focus on the case when the state $x$ is fully observable, but Section~\ref{implementation} discusses methods for ensuring robustness to imperfect state measurements or observation-feedback control.

Given an initial state $x_0 \in \X_0 \subseteq \X$ at time $t=0$, the closed-loop system's behavior can be described via its trajectory map $x(t) = \xi_\pi(x_0, t): \X_0 \times \R^+ \mapsto \X$, which we associate with the controller $\pi$ used to close the feedback loop. When designing a controller, it is important to ensure that the system's behavior achieves objectives such as stability, safety, or contraction. In the rest of this section, we will define each of these objectives in turn and introduce certificates that can be use to prove that a controller achieves these objectives.

\subsection{Stability and Lyapunov certificates}

Stability can be defined in a number of increasingly strict senses \cite{khalil}. A point $x_g \in \mathcal{X}$ is said to be:
\begin{enumerate}[label=a)]
    \item Stable in the sense of Lyapunov (i.s.L.) if, for an appropriate norm, for every $\epsilon > 0$ there exists a $\delta > 0$ such that for all $t_2 \geq t_1 \geq 0$:
    \begin{align}
        \norm{x(t_1) - x_g} \leq \delta \implies \norm{\xi_\pi(x_0, t_2) - x_g} \leq \epsilon \label{stabe_isl_def}
    \end{align}
    
    \item Asymptotically stable if the system is stable i.s.L. and
    \begin{align}
        \lim_{t \to \infty}\norm{\xi_\pi(x_0, t) - x_g} = 0 \quad \forall \quad x_0 \in \X_0 \label{stabe_asy_def}
    \end{align}
    
    \item Exponentially stable if there exist constants $C, \lambda > 0$ such that
    \begin{align}
        \norm{\xi_\pi(x_0, t) - x_g} \leq C \norm{x_0 - x_g}e^{-\lambda t} \quad \forall \quad x_0 \in \X_0 \label{stabe_exp_def}
    \end{align}
\end{enumerate}

The concept of stability can also be extended to the case where the goal is a set $\Xg \subseteq \X$ rather than a point, in which case the norm distance between points in the above hierarchy is replaced with distance between a point and $\Xg$ (see Definition 4.10 in~\cite{Wassim}).

To prove that a closed-loop system is stable about a point, we can turn to one of the most widely known types of certificate: the Lyapunov function. A continuously differentiable function $V: \X \mapsto \R$ is a Lyapunov function if
\begin{subequations}
\begin{align}
    V(x_g) &= 0 \label{lyap_zero} \\
    V(x) &> 0\quad \forall x \in \X \setminus \set{x_g} \label{lyap_PD} \\
    \der{V}{t} &\leq 0\quad \forall x \in \X \label{lyap_decrease_semi}
\end{align}
\end{subequations}
Where $\der{V}{t} = \nabla V(x) f_{c}(x)$ is the Lie derivative of $V$ along the closed-loop dynamics $f_{c}$ (often denoted $L_{f_c}V(x)$). If a function satisfying these conditions can be found, then it serves to certify the stability of $x_g$ via the following theorems.

\begin{theorem}
[Theorem 4.1 in \cite{khalil}] If $V$ is a Lyapunov function (i.e. $V(x_g) = 0$, $V(x) > 0\ \forall\ x \in \X \setminus \set{x_g}$, and $\der{V}{t} \leq 0\ \forall\ x \in \X$) and $f_c(x_g) = 0$, then the system has a stable i.s.L. equilibrium at $x_g$ (with $\X$ forming the region of attraction). Moreover, if $\der{V}{t} < 0$ for all $x \in \X \setminus \set{x_g}$, then the system has an asymptotically stable equilibrium at $x_g$.
\end{theorem}

The proof can be found in most controls textbooks and is omitted here, but the primary insights are 1) sub-level sets of $V$ are forward invariant (i.e. once the system enters a sub-level set of $V$, it will remain inside that set for all future time), proving stability i.s.L., and 2) if $V$ is monotonically decreasing and bounded below, then it must eventually approach its minimum value at $0$. Intuitively, it is useful to think of $V$ as a generalized energy --- if the system is strictly dissipative then it will eventually come to rest. We can also certify exponential stability by adding additional conditions to $V$.

\begin{theorem}\label{lyap_theorem}
[Theorem 4.10 in \cite{khalil}] If $V$ is a Lyapunov function, $f(x_g) = 0$, and there exist positive constants $k_1, k_2, k_3, a$ such that
\begin{subequations}
\begin{align}
    k_1 \norm{x - x_g}^a &\leq V \leq k_2 \norm{x - x_g}^a  \label{V_exp_norm}\\
    \der{V}{t} &\leq -k_3 \norm{x - x_g}^a \label{V_exp_decrease}
\end{align}
\end{subequations}
then $x_g$ is exponentially stable.
\end{theorem}

The proof can be found in \cite{khalil}; the main insight is that condition~\eqref{V_exp_decrease} ensures that $V$ decays exponentially to zero, while condition~\eqref{V_exp_norm} ensures that $V \to 0$ implies $||x - x_g|| \to 0$. Other forms of Theorem~\ref{lyap_theorem} exist in the literature~\cite{dawson2021safe}, typically replacing~\eqref{V_exp_decrease} with $L_{f_c}V \leq -k_3 V$, but these modifications do not change the essential theory.

\begin{figure}[h]
    \centering
    \includegraphics[width=0.6\linewidth]{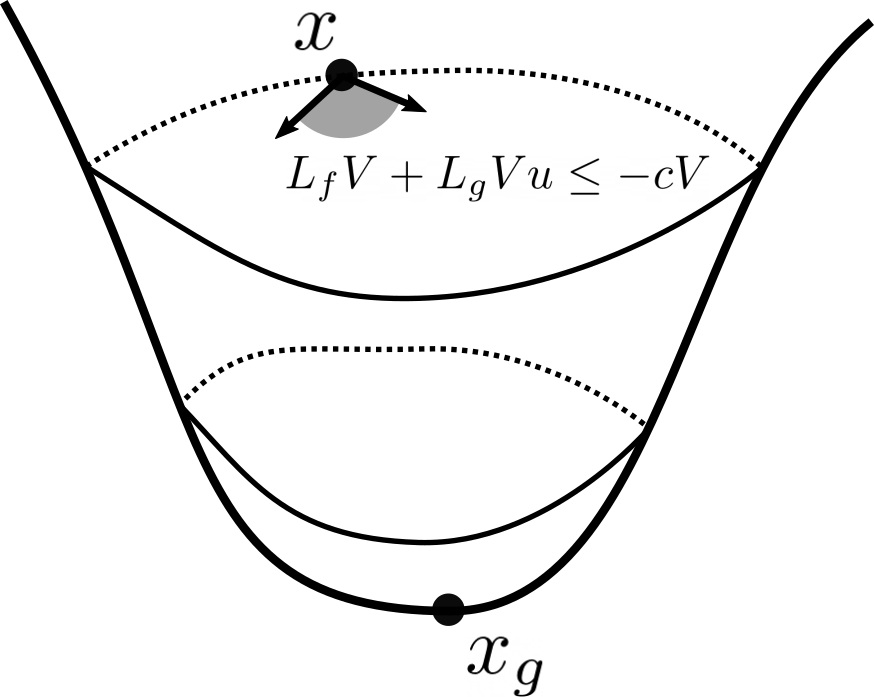}
    \caption{A control Lyapunov function can be used to filter a potentially-unstable control input by solving a quadratic program. The contours illustrate how $V$ varies with $x$, and the gray region indicates the potential values of $\dot{x}$ for different values of $u \in K_{ESCLF}$.}
    \label{fig:clf_illustration}
\end{figure}

In addition to certifying the stability of a closed-loop system, Lyapunov functions can also be applied to certify the \textit{stabilizability} of open-loop systems (i.e. prove that there exists some controller that makes the closed-loop system stable). Restricting our view to a control affine system, a Control Lyapunov Function (CLF) $V: \X \mapsto \R$ certifies asymptotic stabilizability about $x_g$ if
\begin{subequations}
\begin{align}
    V(x_g) &= 0 \label{clf_condition_goal}\\
    V(x) &> 0\quad \forall x \in \X \setminus \set{x_g} \\
    \inf_{u \in \mathcal{U}} \left[L_f V(x) + L_g V(x) u \right] &\leq 0\quad \forall x \in \X \label{clf_condition}
\end{align}
\end{subequations}
where $L_f V$ and $L_g V$ denote the Lie derivatives of $V$ along $f$ and $g$, respectively. A similar definition exists for exponentially stabilizing CLFs (ESCLF)~\cite{ames_esclf}, where the last condition is replaced with
\begin{align}
    \inf_{u \in \mathcal{U}} \left[L_f V(x) + L_g V(x) u + c V(x)\right] &\leq 0\quad \forall x \in \X \label{esclf_condition}
\end{align}
for some positive $c$. Once found, a CLF (or ESCLF) defines a set of admissible control inputs for each state:
\begin{align}
    K_{CLF} &= \set{u : L_f V(x) + L_g V(x) u \leq 0} \\
    K_{ESCLF} &= \set{u : L_f V(x) + L_g V(x) u + c V(x) \leq 0}
\end{align}
Any Lipschitz policy $\pi$ that chooses control inputs from these sets will necessarily stabilize the system. Since these conditions are affine in $u$, a common choice is a quadratic program (QP) that finds the smallest-magnitude control such that $u \in K_{CLF}$ or $u \in K_{ESCLF}$ \cite{ames_esclf}, for example:
\begin{subequations}
\begin{align}
    \min_{u \in \mathcal{U}} &\ ||u||^2 \label{clf_qp_controller}\\
    \text{s.t.} &\ L_f V(x) + L_g V(x) u \leq -cV(x)
\end{align}
\end{subequations}
This QP can also be used to filter a potentially-unstable reference policy $\pi_r$ by replacing the objective with $||u - \pi_r||^2$, in which case the QP solves for the value of $u$ closest to the reference that still ensures stability, as shown in Fig.~\ref{fig:clf_illustration}.

Now that we have derived a control policy from the solution of a constrained optimization problem, it is important to discuss the continuity of that policy. An important result in the theory of CLFs (also applicable to CBFs, which we discuss next) is that the controller~\eqref{clf_qp_controller} is Lipschitz continuous (Section 4.2 in~\cite{Freeman1996} and~\cite{ames_esclf}) and has a closed-form solution so long as $\mathcal{U}$ is convex \cite{ames_esclf}.

\subsection{Safety and barrier certificates}

Given an unsafe set $\Xu \subseteq \X$ that does not intersect the set of initial conditions $\X_0$, a system is safe if it will never enter the unsafe region if started within $\X_0$. That is, it will be safe if all trajectories satisfy:
\begin{align}
    x_0 \in \X_0 \implies \xi_\pi(x_0, t) \notin \Xu \quad \forall \quad t \geq 0 \label{safe_def}
\end{align}
To prove this property, we would need to prove that some set (containing $\X_0$ but not intersecting $\Xu$) is forward invariant. We can do this using a certificate known as a barrier function.

Concretely, consider a compact set $\mathcal{C}$, defined as the zero sub-level set of a barrier function $h: \X \mapsto \R$ (i.e. $\mathcal{C} = \set{x : h(x) \leq 0}$). If $h$ satisfies certain properties (given in Theorem~\ref{barrier_thm}), then $\mathcal{C}$ will be forward invariant. Note that some references reverse the sign of $h$ to be positive on the invariant set; however, we choose this convention to illustrate the parallels between barrier and Lyapunov certificates. In particular, it is often useful to know that every Lyapunov function implies a family of barrier functions defined via its sub-level sets.

\begin{theorem}\label{barrier_thm}
[Proposition 1 in \cite{ames_cbf}] If there exists a strictly increasing scalar function $\alpha: \R \to \R$ such that $\alpha(0) = 0$ (i.e. an extended class-$\mathcal{K}$ function) and
\begin{align}
    \der{h}{t} \leq -\alpha(h(x)) \quad \forall x \in \X \label{bf_decrease}
\end{align}
then $h$ is a barrier function and $\mathcal{C}$ is forward invariant for the closed-loop system $\dot{x} = f_c(x)$.
\end{theorem}

A common choice for $\alpha(h)$ is simply $\gamma h$ for some positive $\gamma$. The proof of this theorem can be found in \cite{ames_cbf}; the main insight is that $h$ is decreasing whenever $x$ is on the boundary of $\mathcal{C}$, where $h(x)=0$, which prevents the system from ever exiting $\mathcal{C}$. In fact, if the system starts outside of $\mathcal{C}$, this condition ensures that it will asymptotically converge to $\mathcal{C}$. A direct consequence of Theorem~\ref{barrier_thm} is that if 
$\X_0 \subseteq \mathcal{C}$ and $\Xu \cap \mathcal{C} = \emptyset$, then the closed-loop system is safe in the sense of the definition in~\eqref{safe_def}.

Barrier functions can be extended to certify the safety of open-loop systems through the use of Control Barrier Functions (CBFs), which directly parallel the CLFs introduced in the previous section. A function $h$ is a CBF if
\begin{align}
    \inf_{u \in \mathcal{U}}\left[ L_f h(x) + L_g h(x) u + \alpha(h(x)) \right] \leq 0 \label{cbf_condition}
\end{align}
As with CLFs, this condition is affine in $u$, allowing a CBF to be used in a QP in a similar way. Usefully, the CBF condition~\eqref{cbf_condition} can be combined with a relaxed version of the CLF condition~\eqref{clf_condition} to yield a single controller that combines safety and stability considerations~\cite{ames_cbf}:
\begin{subequations}
\begin{align}
    \min_{u \in \mathcal{U}} &\ ||u||^2 + k r \label{clf_cbf_qp_controller}\\
    \text{s.t.} &\ L_f V(x) + L_g V(x) u + cV(x) \leq r \label{clf_cbf_qp_controller:clf}\\
    & L_f h(x) + L_g h(x) u + \alpha(h(x)) \leq 0 \label{clf_cbf_qp_controller:cbf} \\
    & r \geq 0
\end{align}
\end{subequations}
In this combined CLF-CBF QP, the constraint~\eqref{clf_cbf_qp_controller:cbf} ensures safety at all times while the stability constraint~\eqref{clf_cbf_qp_controller:clf} is relaxed by some variable amount $r$. This relaxation allows the system to temporarily cease progress towards the goal in order to remain safe, and $k > 0$ is a tunable parameter governing the trade-off between control effort and relaxation of the CLF condition. It is important to note that this combined QP can suffer from deadlock when there is no safe direction that moves closer to the goal, although there are proposals for switched controllers~\cite{dawson22perception} and unified Lyapunov-barrier certificates~\cite{dawson2021safe} that alleviate this concern.

\subsection{Contraction metric certificates}

In our previous discussion of stability, we assumed that the control system designer is interested in stabilizing the system about a fixed point; however, in many applications we are more interested in trajectory tracking rather than stabilization. Of course, a fixed-point-tracking controller can be used to track a trajectory if that trajectory varies slowly enough, but in order to track more general trajectories we need a richer specification than simply stabilizing a fixed point.

In this context, the designer wishes to find a controller so that the trajectories $x(t)$ of the closed-loop system will converge to some desired trajectory, which we represent as a sequence of states and control inputs $x^*(t)$ and $u^*(t)$:
\begin{align}
    \norm{x(t) - x^*(t)} \leq R e^{-\lambda t}\norm{x(0) - x^*(0)}\ \forall\ t\geq 0 \label{distance_shrinking}
\end{align}
for some $R \geq 1$, $\lambda > 0$. Intuitively, we can think of this as a requirement that the distance between the system's trajectory and the reference we wish to track shrinks exponentially (with some potential overshoot allowed by $R$). Contraction theory~\cite{contraction, manchester_ccm} allows us to formalize this requirement.

Just as a Lyapunov function defines an energy-like quantity that decays along trajectories (proving stability), contraction theory defines a measure of distance between neighboring trajectories (a \textit{contraction metric}) that decreases exponentially over time~\cite{contraction}. Let the metric $M(x): \X \mapsto \R^{n \times n}$ be a function returning a positive-definite matrix (which we denote $M \succ 0$). This matrix defines a metric (a generalized distance) on virtual displacements between trajectories $\delta x^T M \delta x$. By showing that this metric is contracting (i.e. the displacement between any two neighboring trajectories is shrinking), a contraction metric allows us to certify that a closed loop system (with some feedback tracking controller) is capable of tracking any feasible trajectory using the following theorem.

\begin{theorem}[Proposition 1 in \cite{Sun2020}]
Let $\dot{M} = \sum_{i=1}^n \pder{M}{x_i} \dot{x}_i$, $A = \pder{f}{x}$, $B = \pder{f}{u}$, and $K = \pder{\pi}{x}$ for some feedback tracking controller $\pi(x, x^* ,u^*)$. Further, let $\rm{sym}(A) = A + A^T$.  Then, if for all $x \in \X$, $x^* \in \X$, and $u^* \in \U$,
\begin{align}
    \dot{M} + \rm{sym}(M(A+BK)) + 2\lambda M \prec 0 \label{contraction_condition}
\end{align}
then equation~\eqref{distance_shrinking} holds with $R$ equal to the square root of the condition number of $M$ and we say the system is contracting with rate $\lambda$.
\end{theorem}

\begin{figure}[h]
    \centering
    \includegraphics[width=0.5\linewidth]{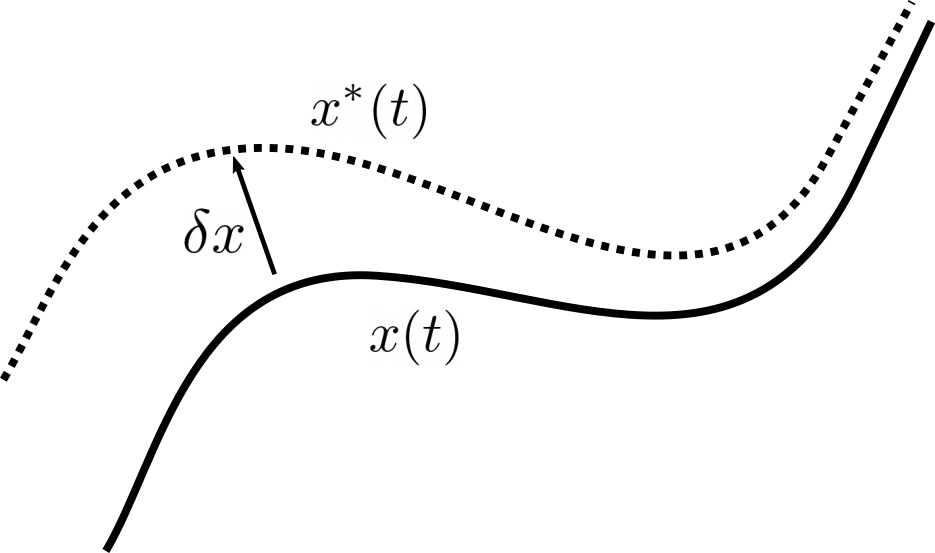}
    \caption{A contraction metric proves that the system will exponentially converge to track any feasible reference trajectory. The error $\norm{x(t) - x^*(t)}$ shrinks exponentially, with transient overshoot limited by the condition number of $M$ and the steady-state tracking error is bounded proportional to the worst-case disturbance.}
    \label{fig:cm_illustration}
\end{figure}

Contraction is a property of the closed-loop system, not of any particular trajectory, and thus the existence of a metric satisfying condition~\eqref{contraction_condition} suffices to prove that a control system is capable of exponentially stabilizing any dynamically feasible nominal trajectory. Additionally, if a system is contracting, then bounded disturbances will produce a bounded worst-case tracking error, and this tracking error will be proportional to the magnitude of the disturbance \cite{Sun2020}. As a result, the metric $M$ provides a certificate that the corresponding tracking controller $\pi$ successfully stabilizes the system and is robust to disturbances, as illustrated in Fig.~\ref{fig:cm_illustration}.

\section{Prior Work on Certificate Synthesis}\label{related-work}

In this section, we discuss prior work on certificate synthesis methods. We separate this discussion into two parts. First, we cover early methods for certificate synthesis, including numerical methods, polynomial optimization, and simulation-guided synthesis. These methods represent an important step in the history of certificate synthesis, but the poor scalability of these early methods, particularly when dealing with nonlinear dynamics, sparked the development of the neural techniques that are the focus of this survey.

Of course, neural certificates are not the only class of methods that have evolved to fill this gap. The second part of this section discusses two parallel lines of work, safe RL and reachability, that are similar in purpose to neural certificate methods but often use different vocabularies and methods. Our aim here is not to provide a comprehensive introduction to these methods, since excellent surveys exist for both Hamilton-Jacobi reachability~\cite{bansal_hji} and safe RL~\cite{brunke_safe_rl}; instead, we aim to to highlight key similarities and differences between these lines of research and situate recent developments in neural certificates in this modern context.

\subsection{Early approaches: numerical solutions and optimization}

Depending on the complexity of the dynamics, several existing techniques may be used to synthesize certificate functions, although there is currently no generally-applicable scalable framework~\cite{Giesl2015}. When the dynamics of the closed-loop system are linear and stable, $\dot{x} = (A - BK)x$ for feedback gains $K$, then a quadratic Lyapunov function will exist of the form $V(x) = x^T P x$ for some symmetric positive definite matrix $P$. In this case, $P$ can be found by solving the continuous Lyapunov equation numerically (most numerical linear algebra packages provide this functionality, including MATLAB~\cite{matlab_cle} and SciPy~\cite{2020SciPy-NMeth}).

For systems with polynomial dynamics, constraints on the sign of the certificate function's derivative can be expressed as constraints that certain polynomials be expressible as sums-of-squares (SoS). Due to the correspondence between fixed-degree SoS polynomials and positive semidefinite matrices (which form a convex cone), certificates encoded in the SoS framework can be synthesized using convex optimization methods. Unfortunately, the computational complexity of these techniques grows exponentially in the degree of polynomials involved and they are limited to systems with polynomial dynamics \cite{Ahmadi2016}. Additionally, SoS optimization is only convex when searching for either a certificate or a controller; the optimization problem becomes non-convex and often encounters numerical issues when searching for a certificate and controller simultaneously \cite{Majumdar2013}. A comprehensive review of SoS methods for control is outside the scope of this review, but the interested reader may see~\cite{Ahmadi2016} for a survey and~\cite{wang_sos_cbf,tan_sos_lyap,Majumdar2013} for relevant examples.

A related technique for certificate synthesis for nonlinear systems is the simulation-guided synthesis~\cite{Kapinski2014}. In this framework, the controls engineer selects a fixed set of basis function $z(x)$ (e.g. monomials up to some degree) and constructs a Lyapunov candidate of the form $V(x) = z^T P z$. The state space is then sampled at a large number of points, and each point is simulated forward to estimate the closed-loop state derivative. After computing $z$ for these fixed points, the Lyapunov and barrier function conditions at those points become linear in the entries of $P$, allowing $P$ to be computed using linear programming. After solving for $P$, the candidate $V$ is  verified using a satisfiability modulo theory (SMT) solver, which a generalizes Boolean satisfiability to include statements involving real numbers~\cite{barrettSatisfiabilityModuloTheories2018}. If the SMT solver finds a violation of the certificate conditions, then a counterexample representing that violation is added to the optimization problem and $P$ is recomputed. This process continues until either the SMT solver verifies that $V$ is a valid certificate or the LP becomes infeasible (in which case we can conclude nothing beyond the fact that our choice of basis is not suitable).

Due to the poor scalability of these methods, particularly with regard to nonlinear or high-dimensional dynamics, they have not seen widespread adoption since their introduction (with the notable exception of SoS methods, although substantial concerns remain about the scalability of SoS methods in practice~\cite{dawson2021safe}). A qualitative summary of various prior methods is shown in Table~\ref{methods_comparison}.

\subsection{Parallel approaches: safe RL and HJ reachability}

It is important to acknowledge that certificate-based control is not the only line of research that deals with the problem of guaranteeing safety and stability for control systems. In particular, safe RL~\cite{brunke_safe_rl} and Hamilton-Jacobi reachability~\cite{bansal_hji} have received a large amount of research interest in recent years. While full coverage of these methods is outside the scope of this survey, we will review the basics of each approach here, with an eye towards highlighting the underlying commonalities and differences between these methods and certificate-based control.

In the broadest possible terms, safe RL is a collection of methods that seek to optimize a control policy to achieve good performance on a specified task (i.e. by maximizing some reward) while respecting a number of constraints on the behavior of the controlled system~\cite{brunke_safe_rl}. There are three classes of safe RL methods in particular that highlight connections to certificate-based control. The first class of methods use an \textit{a priori} given certificate to constrain the actions that the RL agent can take~\cite{Cheng2019,berkenkamp17saferl,Li2019}, possibly with some learning of model uncertainty as the controller explores~\cite{Cheng2019,berkenkamp17saferl}. These methods highlight the complementary nature of certificate-based control and traditional RL; certificates enforce safety and stability constraints while the RL agent tries to achieve good performance on non-safety-critical metrics. Of course, these methods require a certificate to be provided \textit{a priori}, requiring a separate certificate synthesis process, and so they are not the focus of this survey. The second class of safe RL methods propose to learn certificates alongside a control policy using standard RL optimization algorithms (e.g. learning a Lyapunov function in~\cite{Chang2021}, a CLF in~\cite{westenbroek_2020}, or a CLF and CBF together in~\cite{WESTENBROEK202119}). We see these methods as examples of the neural-certificates approach, with the main difference between these and other neural certificate-learning methods being the choice of optimization algorithm (there is also a minor vocabulary difference, as the safe RL community often considers a ``safety index'' function that acts similarly to a barrier function~\cite{ma22_joint_synthesis_rl}). The final class of safe RL methods attempts to derive certificate functions directly from the structure of the constrained partially-observable Markov decision process underlying the safe control problem~\cite{chow2018lyap_rl,chow2019lyapunovbased}; these methods are exciting, as they suggest the future possibility of extending control-theoretic certificates to handle a more general class of constraints (as we discuss in Section~\ref{future_work}).

In addition to the body of safe RL literature, there are also a class of methods for safe control synthesis and verification based on Hamilton-Jacobi (HJ) reachability analysis. HJ methods, which are sometimes also referred to as Hamilton-Jacobi-Isaacs (HJI) methods to emphasize the connection to game theory, typically frame the safe control problem as a two-player zero-sum game between the controller and an adversary. In this game, the controller seeks to maximize some safety index (making the system as safe as possible) while the adversary seeks to minimize this index. Both a value function (indicating how unsafe the system will be if it starts at a given state) and a control policy that maximizes the safety of the system are found by approximately solving an HJI partial differential equation under the assumption of bounded actions by the adversary~\cite{bansal_hji}. There are fundamental connections between the HJ value function and certificate functions, since the super- and sub-level sets of HJ value functions and Lyapunov functions, respectively, certify the forward invariance of a safe region. The primary difference between traditional HJ and certificates is that the controller found by solving an HJ reachability problem is \textit{optimally safe} in that it always seeks to maximize the safety of the system, and can thus be conservative, but recent work in linking HJ reachability with barrier functions~\cite{choi21valuebarrier} have reduced this conservatism. Historically, HJ methods have been limited by the complexity of solving the HJI PDE in high-dimensional state spaces (which usually requires covering the state space with a discrete mesh of sufficient resolution before iteratively solving the PDE,~\cite{bansal_hji}). However, there has been an exciting trend towards using neural networks to approximate solutions of HJ reachability problems~\cite{bansal_deepreach,rubies_royo_classification_hji}, paralleling the use of neural networks for general certificate synthesis, substantially reducing the amount of computation required for HJ analysis. Based on these developments, we anticipate that HJ reachability methods and neural certificates will continue to converge over the coming years, with each community applying insights from the other to develop more scalable and flexible algorithms.

\section{Learning Neural Certificates}\label{learning}

All of the certificates discussed above allow the controls engineer to prove that her controller design is sound (or, in the case of CLFs and CBFs, derive a sound controller directly from the certificate). However, these methods share a common drawback: there has historically been no general, scalable method for finding these certificates \cite{Giesl2015}. In this section, we will discuss how we can apply \textit{self-supervised learning} to learn these certificates using neural networks (so-called because the algorithm requires no human labeling to supervise its learning; it can generate its own labels).

First, we will discuss how a certificate can be found independent of any controller. This case is useful when a controller is known and the designer simply wishes to find a certificate for the closed-loop system, or when the certificate itself can be used to derive a controller implicitly (as with a CLF or CBF). Of course, in some applications we wish to learn an explicit control policy as well; we will discuss this case shortly, but it is instructive to begin with the certificate-only case.

Generally speaking, the search for a certificate can be seen as an optimization over a space of continuously differentiable functions $\mathcal{V}$. If we consider a generic certificate $V: \X \mapsto R^q$ (where $q=1$ for Lyapunov and barrier certificates and $q=n^2$ for contraction metrics), then the search for a generic certificate can be encoded as an optimization problem \cite{Boffi2020}:
\begin{align}
    \text{find}_{V \in \mathcal{V}}\ \text{s.t.}\ c_i\pn{x, V} \leq 0 \quad \forall x \in X \quad i\in \mathcal{I} \label{eq:generic_certificate_search}
\end{align}
where $\mathcal{I}$ is the set of conditions $c_i$ that must hold for this particular certificate (with slight abuse of notation for converting between equality constraints and inequalities). For example, to search for a Lyapunov function, inequalities~\eqref{lyap_zero}, \eqref{lyap_PD}, and~\eqref{lyap_decrease_semi}  must hold, creating the optimization problem
\begin{subequations}
\begin{align}
    \text{find}_{V \in \mathcal{V}}\ \text{s.t.}\ V(x_g) &= 0 \\
    V(x) &\geq 0 \quad \forall x \in \X \setminus \set{x_g} \\
    \der{V}{t} &\leq 0 \quad \forall x \in \X
\end{align}
\end{subequations}

Different approaches to certificate synthesis can be organized according to how they approach this optimization program. For example, the SoS and simulation-guided synthesis methods discussed in Section~\ref{related-work} restrict $\mathcal{V}$ to the span of a chosen set of basis functions (e.g. polynomials of fixed degree), then solve~\eqref{eq:generic_certificate_search} as a convex (or bi-level convex) program. The convexity imposed by a particular choice of $\mathcal{V}$ (e.g. the set of polynomials up to a certain degree) makes solving \eqref{eq:generic_certificate_search} more efficient but can be overly restrictive, resulting in \eqref{eq:generic_certificate_search} becoming infeasible if $\mathcal{V}$ is not rich enough.

The neural certificate framework can be seen as a natural extension of this approach by searching over a much richer function space. To avoid the limitations imposed by the choice of function space, $\mathcal{V}$ is represented as the space of neural networks of particular depth and width (chosen as hyper-parameters by the user). By increasing the size of these networks, this representation can approximate any continuous function \cite{Funahashi1989,Barron1994}, alleviating limitations due to the choice of function space. Although this same universal approximation property also applies to polynomials (i.e. a polynomial of large enough degree can also approximate any continuous function on some domain), in practice the scaling for neural networks is much more favorable (the number of parameters in the input layer of a neural network grows with $O(nd)$ with $n$ input dimensions and $d$ dimensions in the next layer, while the number of monomials in a $d$-degree polynomial in $n$ dimensions grows with $O(n^d)$).

To adapt~\eqref{eq:generic_certificate_search} to be solved using a neural representation for $V$, we must make two modifications to account for the fact that neural networks lend themselves best to unconstrained optimization problems (solved via stochastic gradient descent) rather than constrained optimization. First, we relax the constraints using a penalty method. Second, instead of imposing constraints universally (e.g. $\forall x \in \X$), we evaluate them at a large, finite set of randomly sampled training points $\set{x_1, \ldots, x_N} \subset \X$. Each of these training points can be automatically labeled with a loss equal to the violation of the certificate conditions at that point, and averaging this loss across the entire training set yields the \textit{empirical certificate loss}:
\begin{align}
    \mathcal{L}_V = \sum_{i \in \mathcal{I}} \frac{\alpha_i}{N}\sum_{j = 1}^N\max(c_i\pn{x_j, V}, 0) \label{eq:relaxed_certificate_search}
\end{align}
with penalty weights $\alpha_i$ that are tuned empirically as hyper-parameters to encourage constraint satisfaction. \textit{Loss} refers to the fact that all constraints are included as scalar penalties to be minimized, and \textit{empirical loss} is distinguished from the \textit{true} loss in that it is evaluated at a finite number of points rather than over the entire state space. This empirical loss can be minimized using stochastic gradient descent with respect to the weights and biases of the neural network used to represent $V$, but it is important to remember that zero empirical loss does not guarantee that the certificate is valid; it provides only statistical evidence of validity that can potentially be supplemented by one of the neural network verification strategies discussed in Section~\ref{certificate_verification}.

This approach provides the foundation for many certificate-based learning works; early examples include~\cite{Richards2018}, and later works include~\cite{kolter_neurips2019_stable_deep,Peruffo2020,Abate2020,abate_fossil,Zhao2020,Srinivasan20,gaby2021lyapunovnet,yin_stability_analysis}. Examples that learn CLF or CBF certificates that imply controllers include~\cite{Tsukamoto2020,Robey2020,Lindemann2020,Chen2020,dawson2021safe,dawson22perception,ROBEY20211,xiao2021barriernet,agrawal_cvxpy}. A related class of works assume that a certificate is given for a nominal system and use supervised learning to adapt that certificate to the uncertain true model~\cite{Taylor2019,Taylor2020,Choi2020,Castaneda2020}.

\subsection{Remarks on certificate learning}

At this point it is important to make two remarks. The first remark concerns the constraints that certificates impose on neural network structure. Most of certificate theory for continuous-time dynamical systems assumes that the certificate is continuously differentiable, so it is important that continuous-time certificates are learned using networks with continuously-differentiable activation functions such as $\tanh$, \texttt{softplus}, or the exponential linear unit (ELU). In the discrete-time case, Lyapunov and barrier certificates are required only to be continuous~\cite{grizzle_2001_dt_lyap}; accordingly, works that focus on the discrete-time case have learned certificates using rectified linear unit (ReLU) activation functions, which are continuous but not differentiable at the origin~\cite{Dai2020}.

Second, it is important to note that although we frame the certificate-synthesis problem as a feasibility problem in~\eqref{eq:generic_certificate_search}, there are several metrics of certificate quality that might be included as an objective for the certificate search. The first metric, applicable to Lyapunov functions, is the size of the region-of-attraction certified by the resulting certificate~\cite{ROBEY20211,Chang2019,Richards2018}. Larger regions of attraction can be encouraged either implicitly (by attempting to maximize the percentage of samples in the state space where the certificate is valid~\cite{Chang2019}) or by explicitly searching for new training examples to expand the region of attraction~\cite{Richards2018}. The second metric, applicable to CLFs and CBFs, is the size of the admissible control set $K(x)$, the maximizing of which will yield a less-restrictive controller. This is particularly relevant when actuator limits are present, in which case the size of the intersection of $K$ and $\mathcal{U}$ should be maximized. This second metric is less well explored in the literature. Either objective could be incorporated into the certificate-learning pipeline by adding an appropriately normalized term to the loss function~\eqref{eq:relaxed_certificate_search}, but care should be taken to not allow the neural certificate to overfit to this term at the expense of violating the certificate conditions.

\subsection{Learning certificates and controllers together}

Learning the certificate alone is useful when a controller is already known and we seek to verify its performance, or in cases when learning a certificate such as a CLF or CBF allows us to derive a controller. When it is necessary to explicitly learn a controller alongside the certificate, we can easily extend this framework to simultaneously search for a control policy $\pi$ from some family of policies $\Pi$ (commonly, $\pi$ is parameterized as a separate neural network).

There are three different contexts in which we might wish to learn a control policy. The first case is behavior cloning, where we know an ``expert'' control policy to use as a basis for the learned controller. For example, we may have a computationally expensive nonlinear MPC policy that we wish to compress into a neural network for online computation~\cite{xiao2021barriernet}, or we may have a full state feedback controller that we wish to clone for a partial state feedback context (\cite{margolis2021jumping} provides a non-certificate-based example of this context). To accomplish this certified behavior cloning, we simply augment the empirical loss in~\eqref{eq:relaxed_certificate_search} with a behavior cloning term that penalizes the difference between $\pi$ and the expert policy (often simply the mean of an appropriate norm):
\begin{align*}
    \mathcal{L} = \mathcal{L}_V + \mathcal{L}_{BC}
\end{align*}
Another interesting application of this case is using a barrier certificate to learn more realistic behavior predictions for pedestrians~\cite{meng21_pedestrian}. In this application, the ``expert'' is not a control policy per se but observations of human actions, and the resulting policy $\pi$ is not used for control but to generate more realistic predictions of the humans' future actions.

The second case where we might wish to learn a controller alongside a certificate is when we do not have an expert reference controller, but we have a reward that we seek to maximize; this leads to certificate-regularized reinforcement learning. \cite{Chang2021} provides an example of reinforcement learning with a Lyapunov function where $\mathcal{L}_V$ is used alongside the reward to update the control policy. This setting brings additional complications, as reinforcement learning algorithms typically assume no prior knowledge of the dynamics, and so the Lie derivative of $V$ must be estimated from observed trajectories of the system rather than computed exactly.

The final case is when there are no requirements (beyond the certificate) placed on the learned control policy. In this case, $\mathcal{L}_V$ is used to self-supervise both the certificate $V$ and control policy $\pi$. The first example of simultaneous certificate/policy learning appeared in~\cite{Chang2019} for a Lyapunov certificate; later works learn barrier functions~\cite{Qin2021} and contraction metrics~\cite{Sun2020} alongside control policies as well, or extend to the case when dynamics are only partially-known~\cite{qin2021_sablas}.

Next, we will make this discussion concrete with examples of learning a Lyapunov function, barrier function, and contraction metric, then discuss the history of certificate learning in the controls, robotics, and machine learning literature. Code for each example can be found at \url{https://github.com/MIT-REALM/neural_clbf}.

\subsection{Example: learning a Lyapunov function}\label{ex_lyap}

Our first example involves learning a CLF for the classic toy example: the inverted pendulum (we will progress to more complex examples shortly). This dynamical system has states $x = [\theta, \dot{\theta}]$ and control inputs $u = [\tau]$. The system is parameterized by its mass $m$, length $L$, damping $b$, and gravitational acceleration $g$, and its dynamics have control-affine form:
\begin{align*}
    \dot{x} &= \mat{\dot{\theta} \\ \frac{g}{L}\sin\theta - \frac{b}{mL^2}\dot{\theta}} + \mat{0 \\ \frac{1}{mL^2}} u
\end{align*}

Our goal in learning a CLF is to stabilize this system about $x_0 = [0, 0]$. To learn the CLF, we parameterize $V$ as a neural network with two hidden layers of 64 units and minimize the following empirical loss over $N=10^4$ training points. In all experiments, we sample the training data uniformly from the state space, which we represent as a hyperrectangle.
\begin{align*}
    \mathcal{L}_{V} &= \lambda_1 V(x_0) \\
    & + \frac{\lambda_2}{N}\sum_{i=1}^N r(x_i) \\
    & + \frac{\lambda_3}{N}\sum_{i=1}^N \max\pn{L_f V(x_i) + L_g V(x_i) u_i, 0} \\
    & + \frac{\lambda_4}{N}\sum_{i=1}^N \max\pn{\frac{V(x_i + \Delta t \dot{x}(x_i, u_i)) - V(x_i)}{\Delta t}, 0}
\end{align*}
where $\lambda_1 = 10$, $\lambda_2 = 10^3$, $\lambda_3 = \lambda_4 = 1$, $[r(x_i), u_i]$ are the solutions to the relaxed CLF QP at state $x_i$ with $c=1$ and $\lambda_5 = 10^2$:
\begin{align*}
    \min_{r, u} &\ ||u||^2 + \lambda_5 r \\
    \text{s.t.} &\ L_f V(x) + L_g V(x) u \leq -cV(x) + r \\
    &\ r \geq 0
\end{align*}

\begin{figure}[h]
    \centering
    \includegraphics[width=0.8\linewidth]{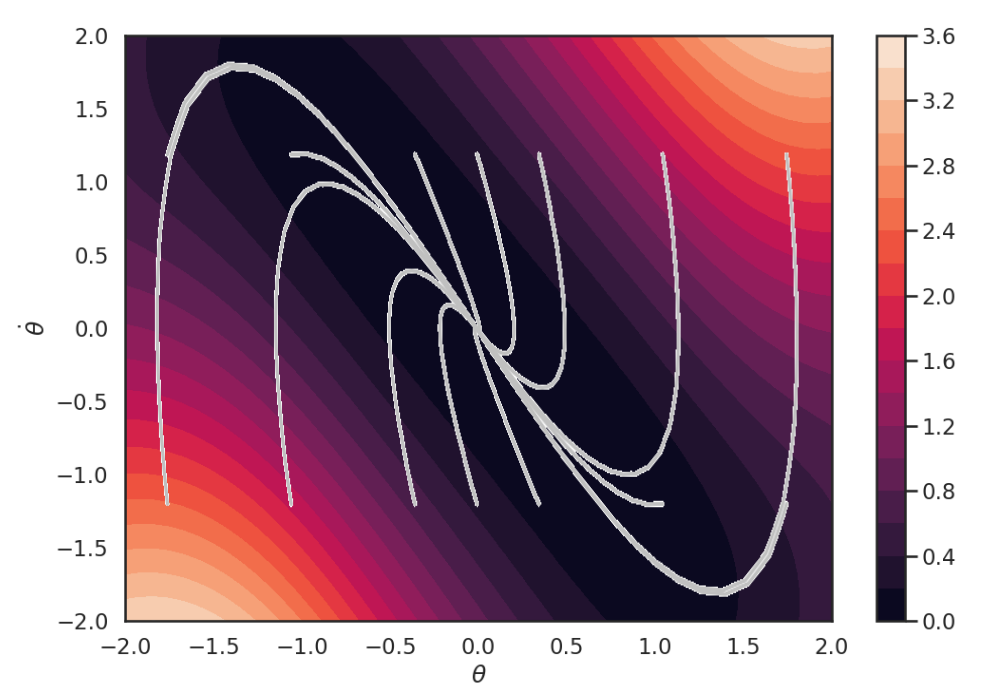}
    \caption{The CLF learned for the inverted pendulum, overlaid with trajectories of the system controlled using the CLF-QP~\eqref{clf_qp_controller}. Lighter colors indicate higher values of $V$; as expected, $V$ decreases to zero at the goal point, and trajectories of the controlled system always move in a direction of decreasing $V$.}
    \label{fig:ip_clf}
\end{figure}

The third and fourth terms of the empirical loss are different approximations of the Lyapunov decrease conditions; strictly speaking, only the first two terms are needed, but additional terms can help the learning process. Note that in this example, we train $V$ to be positive definite, but it is also possible to ensure positive-semidefiniteness by construction by learning some function $\omega(x) : \R \mapsto \R^{n_h}$ as a neural network with $n_h$ output units, then taking $V(x) = \omega(x)^T\omega(x)$ (this construction is helpful for more complicated examples, but the first loss term is still required to encourage a minimum at $x_0$). Code for this example can be found in \texttt{training/train\_inverted\_pendulum.py}. The results of learning a CLF for this example are shown in Fig.~\ref{fig:ip_clf}; the CLF-derived controller successfully stabilizes this simple dynamical system. In the next section, we will set our sights higher with a more complicated example of learning a CBF.

\subsection{Example: learning a CBF}\label{ex_cbf}

To demonstrate learning a CBF, we use the spacecraft rendezvous problem from~\cite{jewison16spacecraft}. In this problem, we need to design a station-keeping controller to keep the ``chaser'' satellite near a ``target,'' respecting both a minimum distance constraint (to avoid collision) and a maximum distance constraint (to enable observation of the target). The state of the chaser is expressed relative to the target using linearized Clohessy-Wiltshire-Hill (CWH) equations, with state $x = [p_x, p_y, p_z, v_x, v_y, v_z]$ and control inputs corresponding to thrust in each direction $u = [u_x, u_y, u_z]$. With mass $m$ and mean-motion orbital parameter $n$, the dynamics are given by
\begin{align*}
    \dot{x} = \mat{
        1 & 0 & 0 & 0 & 0 & 0 \\
        0 & 1 & 0 & 0 & 0 & 0 \\
        0 & 0 & 1 & 0 & 0 & 0 \\
        3n^2 & 0 & 0 & 0 & 2n & 0 \\
        0 & 0 & 0 & -2n & 0 & 0 \\
        0 & 0 & 0-n^2 & 0 & 0 & 0
    } x + \mat{0 & 0 & 0 \\ 0 & 0 & 0 \\ 0 & 0 & 0 \\ 1 & 0 & 0 \\ 0 & 1 & 0 \\ 0 & 0 & 1} u
\end{align*}

The station-keeping constraints define the unsafe states $\Xu = \set{x : 0.25 \geq r \leq 1.5}$ with $r = \sqrt{p_x^2 + p_y^2 + p_z^2}$. To learn a CBF for these constraints, we take a similar approach as in the previous example and define $h$ as a neural network with 2 hidden layers of 256 units each, sample $N=10^5$ points from the state space (labeled according to whether or not they belong to the unsafe or safe sets), and minimize the empirical loss
\begin{align*}
    \mathcal{L}_{V} &= \frac{\lambda_1}{N_{safe}}\sum_{i=1}^{N_{safe}} \max\pn{h(x_i), 0} \\
    & + \frac{\lambda_2}{N_{unsafe}}\sum_{i=1}^{N_{unsafe}} \max\pn{-h(x_i), 0} \\
    & + \frac{\lambda_3}{N}\sum_{i=1}^N r(x_i)
\end{align*}
where $\lambda_1 = \lambda_2 = 100$, $\lambda_3 = 1$, and $r(x_i)$ is found by solving the CBF QP at state $x_i$ with nominal control input $u_0$ (found using LQR), $c = 0.1$, and $\lambda_5 = 10^4$:
\begin{align*}
    \min_{r, u} &\ ||u - u_0||^2 + \lambda_4 r\\
    \text{s.t.} &\ L_f h(x) + L_g h(x) u \leq -ch(x) + r \\
    &\ r \geq 0
\end{align*}

\begin{figure}[h]
    \centering
    \includegraphics[width=0.8\linewidth]{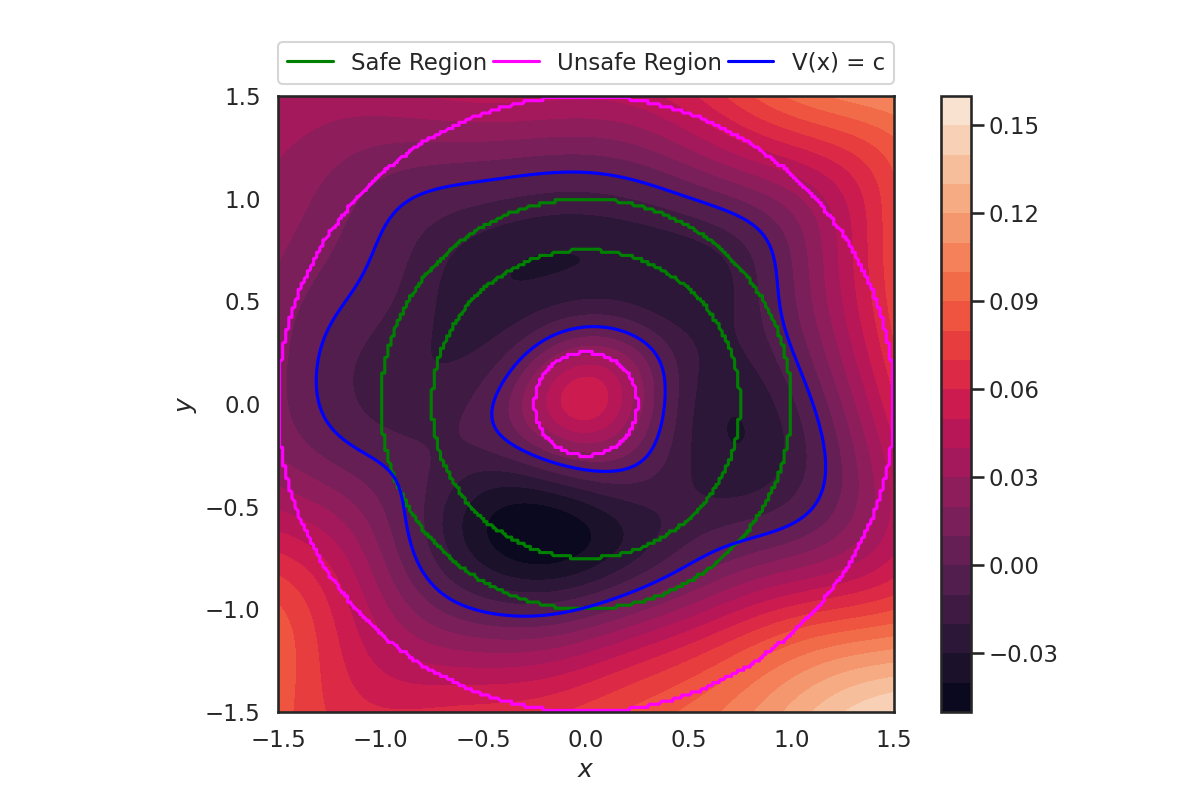}
    \caption{A 2D cross-section of the CBF learned for the satellite station-keeping task, showing how the learned barrier function successfully separates the safe and unsafe sets.}
    \label{fig:satellite_cbf}
\end{figure}

\begin{figure}[h]
    \centering
    \includegraphics[width=0.8\linewidth]{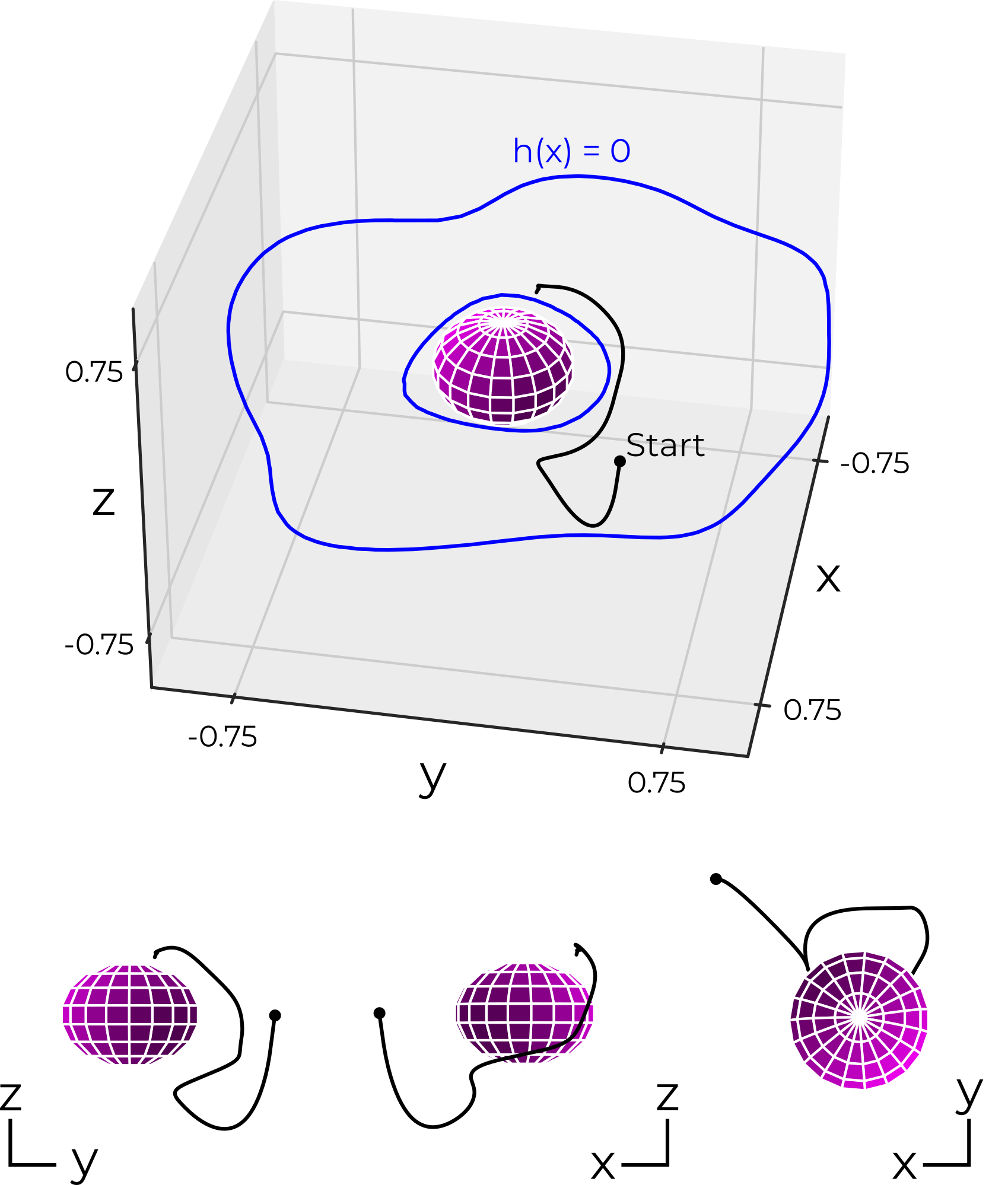}
    \caption{Trajectory followed by chaser satellite subject to station-keeping CBF. The chaser starts by dropping below the target, then circles up to a fixed point, respecting the station-keeping constraints at all times. The blue lines show a 2D cross-section of the zero-level set of the CBF $h$ in the $xy$ plane.}
    \label{fig:satellite_traj}
\end{figure}

Here, the first and second terms of the empirical loss enforce the boundary conditions, training the $h$ network so that its 0-level set segments the safe and unsafe states. The third term is designed to train $h$ so that the corresponding CBF QP is feasible. Code for this example can be found in \texttt{training/train\_linear\_satellite.py}. The learned barrier function is shown in Fig.~\ref{fig:satellite_cbf}, and an example trajectory for the chaser controlled using the CBF QP is shown in Fig.~\ref{fig:satellite_traj}.

\subsection{Example: learning a contraction metric}\label{ex_cm}

Our third example demonstrates learning a contraction metric alongside a control policy. For this example, we consider the problem of designing a nonlinear trajectory-tracking control policy to run on an autonomous ground vehicle. To motivate this problem, we look at the case where the ground vehicle has limited computational resources on-board, so running computationally expensive control policies such as nonlinear MPC may be infeasible. Instead of running this expensive policy on-board, we can use a neural network to learn a computationally inexpensive clone of the MPC policy, and we also learn a contraction metric to ensure that the cloned policy is sound.

Concretely, we consider a simple nonlinear ground vehicle model with states $x = [p_x, p_y, \theta]$ for the 2D position and heading and control inputs $u = [v, \omega]$ for the linear and angular velocities. The system dynamics are control-affine:
\begin{align*}
    \dot{x} = \mat{
        \cos\theta & 0 \\ \sin\theta & 0 \\ 0 & 1
    }u
\end{align*}

We construct an expert control policy by solving a nonlinear MPC problem using CasADi~\cite{andersson_gillis_horn_rawlings_diehl_2018}. Using this expert policy, we construct a training dataset of 100 random reference trajectories of length \SI{10}{s} each (simulated at \SI{100}{Hz} with a \SI{10}{Hz} zero-order hold control signal; data points were sampled at \SI{10}{Hz}). Expert demonstrations are augmented with a small amount of noise to widen the distribution of training data. We represent the contraction metric $M(x)$ and control policy $u(x, x^*, u^*)$ as neural networks with 2 hidden layers of 32 neurons, and construct the metric $M = A + A^T$ where $A$ is the output of the neural network to ensure that the contraction metric is symmetric (other parameterizations exist that ensure that the eigenvalues of $M$ are lower-bounded by construction~\cite{Sun2020}, but we use this construction for simplicity). We also set $u(x, x^*, u^*) = u^* + \pi(x, x - x^*) - \pi(x, 0)$ to ensure that $u(x, x^*, u^*) = u^*$ when $x = x^*$. We simultaneously optimize the parameters of the metric and policy networks to minimize the empirical loss over $N_{train} = 10^4$ tuples $(x, x^*, u^*, u_{expert})$ sampled uniformly from a user-specified hyper-rectangle:
\begin{align*}
    \mathcal{L} &= \mathcal{L}_{M} + \mathcal{L}_{u} \\
    \mathcal{L}_{M} &= \frac{1}{N_{train}}\sum_i [M - \underbar{m} I]_{PD} + \frac{1}{N_{train}}\sum_i[\bar{m}I - M]_{PD} \\
    &\quad + \frac{1}{N_{train}}\sum_i[\dot{M} + \text{sym}(M(A + BK)) + 2\lambda M]_{ND} \\
    \mathcal{L}_{u} &= \frac{1}{N_{train}}\sum_i ||u(x_i, x^*_i, u^*_i) - u_{expert, i}||^2
\end{align*}
where $[\circ]_{PD}$ and $[\circ]_{ND}$ are hinge losses encouraging positive- and negative-semi-definiteness, respectively. Example code for this example is provided in \texttt{training/contraction/train\_cm.py}; note that the structure of this code is different from that in the previous two examples since contraction metrics must be trained using trajectory data rather than randomly sampled points. Additional examples of contraction metric training can be found in the code accompanying~\cite{Sun2020}: \url{https://github.com/sundw2014/C3M}.

\begin{figure}[h]
    \centering
    \includegraphics[width=0.8\linewidth]{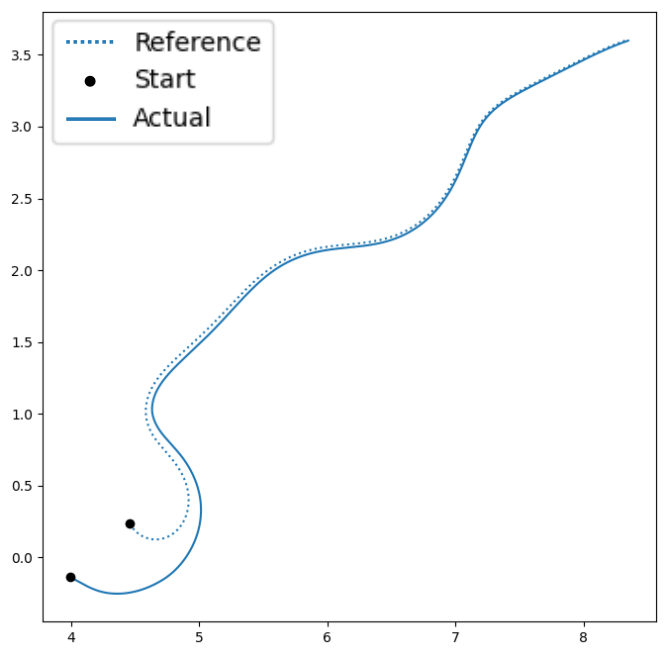}
    \caption{The cloned policy (learned alongside a contraction metric) successfully tracks a previously-unseen reference trajectory. The reference trajectory was generated using a random combination of sinusoidal control inputs.}
    \label{fig:cm_example_traj}
\end{figure}

\begin{figure}[h]
    \centering
    \includegraphics[width=0.8\linewidth]{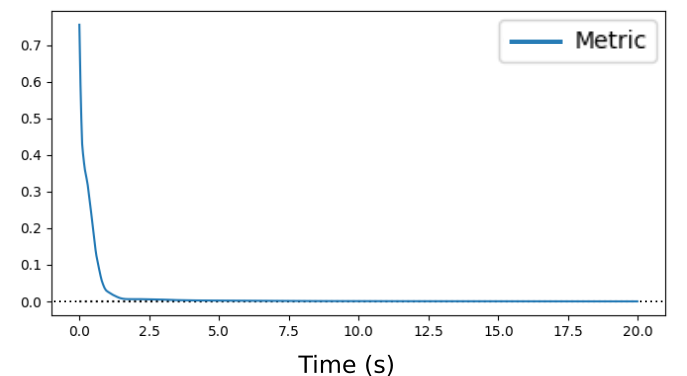}
    \caption{The metric value $(x-x^*)^TM(x-x^*)$ over time as the learned control policy tracks the reference policy. As expected, the contraction metric value decreases exponentially as the system converges to the reference trajectory.}
    \label{fig:cm_example_metric}
\end{figure}

An example of the trained controller tracking a (previously-unseen) random reference trajectory is shown in Fig.~\ref{fig:cm_example_traj}, and the value of the learned contraction metric over time is shown to decrease exponentially in Fig~\ref{fig:cm_example_metric}. Note that even though the neural network control policy only receives the reference state and input at the current instant in time (the original MPC policy receives a 1-second window of future reference states and inputs), it is able to successfully track a trajectory that was not part of the training dataset.

\subsection{History of certificate learning}

The earliest proposal for using a neural network for Lyapunov function synthesis comes from Prokhorov's 1994 paper~\cite{Prokhorov1994}, but this technique relies on neural networks as a model of computation rather than as a function approximator, preventing the approach from scaling. Serpen~\cite{Serpen2005} in 2005 first framed the neural network certificate-representation problem as an optimization problem analogous to~\eqref{eq:relaxed_certificate_search}. However, neither of these works include a demonstration of their learned certificates. Noroozi \textit{et al.}~\cite{Noroozi2008} demonstrated learning a neural network approximation of a Lyapunov function for simple two- and three-dimensional nonlinear dynamical systems in 2008, while Petridis and Petridis~\cite{Petridis2006} in 2006 employ a neural certificate trained via a genetic algorithm to verify the stability of a recurrent neural network. These early works are typically restricted to Lyapunov certificates and involved relatively small dynamical systems.

It is not until after the explosive growth in interest and computational power applied to neural networks in the mid-2010s that neural certificates were rediscovered in the context of robotics and control. One of the first practical implementations came in 2018, where Richards, Berkenkamp, and Krause~\cite{Richards2018} frame certificate synthesis as a classification problem, with the goal of finding the largest region of attraction (RoA) certified by a neural network certificate (they search only for a certificate and assume a controller is given). This approach iteratively grows the approximate certified RoA outwards from a fixed point, sampling from points near the boundary of the RoA to expand its training dataset. The Lyapunov conditions are only enforced for points that lie within the current estimate of the RoA (as determined by simulating forward from those points), allowing this method to estimate the RoA for systems that are only locally stable. Richards \textit{et al.} demonstrate their method on a feedback-stabilized inverted pendulum in simulation, where it outperforms other methods for estimating the RoA (particularly, local linearization and SoS).

Since this initial work, several other authors have also addressed this issue of finding a certificate for a fixed, pre-defined controller. \cite{Ahmed2020,Abate2020} incorporate an SMT solver to verify the learned Lyapunov certificates and provide counterexamples for training, and~\cite{Peruffo2020} takes a similar approach to learning barrier functions (the authors of~\cite{Ahmed2020,Abate2020,Peruffo2020} later unified these methods in a single software framework in~\cite{abate_fossil}). \cite{Singh2020} learns a contraction metric in the case when the control policy is known but applies this technique for system identification rather than controller verification (the contraction metric constrains the learned dynamics model to be stabilizable).

The first work to propose jointly optimizing for a control policy and Lyapunov certificate is that of Chang, Roohi, and Gao in 2019~\cite{Chang2019}. This work employs an SMT-based learner-verifier architecture using the dReal SMT solver, and alternates between optimizing the empirical loss~\eqref{eq:relaxed_certificate_search} and searching for counterexamples to guide training. Chang \textit{et al.} demonstrate that jointly optimizing for a controller and a certificate yields improved performance, in the sense of a larger region of attraction, than either optimizing for a fixed controller (using LQR) or using SoS techniques.

Since Chang \textit{et al.}'s work, other authors have expanded this joint control-certificate learning approach to other certificate functions. For instance, \cite{Qin2021} jointly learns a barrier function and controller, while \cite{Robey2020,Lindemann2020,Chen2020,dawson2021safe,dawson22perception,ROBEY20211} learn control Lyapunov and control barrier functions (implicitly combining the certificate and control policy). Other authors, notably Tsukamoto, Chung, and Slotine~\cite{Tsukamoto2020,Tsukamoto2020b} have extended the neural certificate approach to contraction metrics (see~\cite{tsukamoto21survey} for an overview of these approaches). In addition to different types of certificate, later works have expanded this framework to more challenging problems in control, such as multi-agent control in~\cite{Qin2021}, black-box dynamical models in~\cite{qin2021_sablas}, the RL context in~\cite{Chang2021}, and the robust case in~\cite{dawson2021safe}. Many of these later works are particularly notable for providing theoretical and algorithmic contributions to address difficulties that arise in practice (such as control from observations), which we discuss next in Section~\ref{implementation}.

\section{Implementation Considerations}\label{implementation}

Successfully deploying a control system on hardware is more challenging than demonstrating the performance of that system in simulation. In hardware, effects such as state estimation error, control frequency and delay, external disturbances, unmodeled dynamics, and actuator limits can degrade the safety and stability of the controller unless suitable steps are taken to mitigate these effects. This section will discuss a number of strategies to mitigate these effects. 

\subsection{Mitigating measurement uncertainty}\label{measurement-uncertainty}

A notable gap between simulation and real hardware is the availability of high-quality state estimates. As a result, it is natural to ask how state estimation errors can affect the guarantees of certificate-based controllers. The most directly relevant work in this vein is that by Dean \textit{et al.} deriving conditions under which barrier functions are robust to measurement errors~\cite{Dean2020a}. In this framework, the authors assume access to a state estimate with bounded error, i.e. $\hat{x} = x + e(x)$ with error bounded by some known function $||e(x)|| \leq \epsilon(x)$. Given this assumption and knowledge of the Lipschitz constants of the barrier function and its Lie derivatives, the authors in \cite{Dean2020a} derive a tightened version of condition~\eqref{cbf_condition} that guarantees control-invariance despite state estimation uncertainty. These conditions can be used derive quantitative requirements on the state-estimation accuracy needed to enable a certificate-driven approach to control.

Although \cite{Dean2020a} provides the theoretical basis for barrier certificates that are robust to errors in state estimation, these ``measurement-robust'' barrier functions are strictly more difficult to synthesize than standard barrier functions (due to the tightening of condition~\eqref{cbf_condition}). We are not aware of any published work learning (or otherwise automatically synthesizing) measurement-robust barrier functions, but we anticipate that future work in this area will bridge this gap, since neural synthesis methods can be easily adapted to these tightened conditions.

\subsection{Observation-feedback control certificates}

In Section~\ref{measurement-uncertainty}, we ask how certificates might be adapted to handle an uncertain state estimate; a natural extension of this question is how certificates might be adapted to the case where the control policy is a function of observations themselves (i.e. observation-feedback or ``pixels-to-torques'' control). This is an active area of research in the certificate-learning setting. Early works in this direction assume that the observations are first converted to state estimates with some bounded error (e.g.~\cite{Dean2020a}, discussed in~\ref{measurement-uncertainty}). Later works have provided proofs-of-concept for certificates defined directly in the space of observations, without recourse to an intermediate state estimate~\cite{dawson22perception}. These works typically rely on some approximate model of how control actions affect future observations. For some observations, we can construct an analytical approximate model, for example by approximating future Lidar observations as affine transformations of previous observations~\cite{dawson22perception}, or modeling force measurements using spring connections to the environment~\cite{dawson_force}. For more complicated observations, particularly image feedback, there has been some work towards approximating future images using generative models~\cite{katz_gan,tong_nerf}, but there is still work to be done to understand the approximation errors of deep generative models and the impact of those errors on the safety of a certificate-based controller.

\subsection{Robustness to disturbance and model uncertainty}\label{robustness}

A common goal when designing control systems is to ensure that disturbances to the nominal system dynamics do not adversely affect the safety or performance of the system. Some disturbances, such as unmodeled aerodynamic effects~\cite{shi_neural_lander}, can be represented as unknown forces added to the known system dynamics, while others, such as uncertain mass or inertia, must be modeled as potentially multiplicative disturbances~\cite{dawson2021safe}. Depending on the prior knowledge of the disturbance, different certificate-based control strategies (each with distinct advantages and drawbacks) may be applied to ensure safety.

When little is known about an additive disturbance beyond its bound, i.e. $\dot{x} = f(x, u) + d$, $d \in \mathcal{D}$ for a known disturbance set $\mathcal{D}$, then learning a contraction metric will guarantee that any bounded disturbance will lead to a bounded tracking error in the worst case. When more information about the structure of disturbance is known, for instance when $\dot{x} = f(x, u, d)$ and $f$ is known to be affine in $d$ for any fixed $(x, u)$ (as is the case for additive disturbance as well as uncertainty in many physical parameters), then robust variants of Lyapunov and barrier functions exist to guarantee safety and stability despite this uncertainty \cite{dawson2021safe}. These robust methods are complementary with the adaptation techniques discussed in Section~\ref{certificate_adaptation}, where we might wish to combine a partially-learned dynamics model with robustness to error in those learned dynamics~\cite{Castaneda2020,Choi2020}.

The most challenging case of model uncertainty is when there is no \textit{a priori} knowledge of the structure (e.g. additive, control-dependent, etc.) or extent of the uncertainty. In these cases, the model uncertainty may be estimated from data; \cite{taylor_21_ccf} discusses this case for control-affine systems. While~\cite{taylor_21_ccf} does not include a method for synthesizing a CLF or CBF certificate in this setting, the authors provide a theoretical discussion on how to derive a control input from these certificates (extending the certificate-based QP controllers~\eqref{clf_qp_controller} and~\eqref{clf_cbf_qp_controller}) and when those controllers are feasible despite model uncertainty.

\subsection{Certificate adaptation}\label{certificate_adaptation}

In Section~\ref{robustness}, we discussed robust certificate techniques that take a worst-case approach to handling model uncertainty. This raises the natural question: instead of trying to be robust to a wide range of possible model errors, could we rely on the nominal model for the bulk of the training process and then transfer the learned certificates to the true model using a smaller amount of data from the real system, thus ``adapting'' the certificate to the true model?

One line of work in certificate adaptation takes inspiration from system identification: if the difference between the nominal and real models can be learned, then certificate-based controllers can adjust for this difference at runtime. For example, if control barrier function $h$ is known for the nominal system (which we write as $\hat{f}$ and $\hat{g}$ to distinguish from $f$ and $g$ for the real control-affine system), then a safe controller can be found by solving the CBF QP~\cite{Taylor2020}:
\begin{align}
    \min_{u \in \mathcal{U}}\ & ||u||^2 \\
    \text{s.t.}\ & L_f h(x) + L_g h(x) u \leq -h(x)
\end{align}
Unfortunately, the constraint of this QP depends on the unknown dynamics. If the nominal model differs from the true dynamics with error terms $\Delta_f(x)$ and $\Delta_g(x)$ (i.e. $\dot{x} = \hat{f}(x) + \hat{g}(x)u + \Delta_f(x) +\Delta_g(x) u$), then this introduces an error term:
\begin{align}
    \min_{u \in \mathcal{U}}\ & ||u||^2 \\
    \text{s.t.}\ & L_{\hat{f}} h(x) + L_{\hat{g}} h(x) u + a(x) + b(x)u \leq -h(x)
\end{align}
where $a = L_{\Delta f(x)}h$ and $b = L_{\Delta g(x)}h$. Choi, Casten\~neda, \textit{et al.} demonstrate that a neural network can be used to learn either the model residue $(\Delta_f, \Delta_u)$ or the Lie derivative residue $(a, b)$ via reinforcement learning~\cite{Choi2020} or Gaussian process regression~\cite{Castaneda2020}. They apply their model-learning adaptation technique to a simulated bipedal walker with 14 state dimensions, demonstrating successful adaptation when the mass and inertia of each robot link are changed~\cite{Choi2020}. In a similar vein, Taylor \textit{et al.} learn $a$ and $b$ using an episodic learning framework for both barrier~\cite{Taylor2020} and Lyapunov~\cite{Taylor2019} certificates applied to a Segway model in simulation.

\subsection{Certificate verification}\label{certificate_verification}

Several methods for verifying learned certificates have been proposed, each with its own advantages and drawbacks. Here, we will discuss the three most common methods for verification: probabilistic methods based on generalization error bounds, Lipschitz arguments, and optimization-based methods.

The first framework for verifying learned certificates, generalization error bounds, is the least computationally expensive but provides relatively soft guarantees. These methods typically sample a large number of points from the state space, check whether the certificate is valid with some margin, and then use this margin to extrapolate (with high probability) to claim that the certificate is valid continuously throughout the state space. These claims are typically based on either statistical learning theory (e.g.~\cite{Boffi2020} or Proposition 2 in~\cite{Qin2021}) or almost Lyapunov theory~\cite{Liu2020} (see ~\cite{Chang2021} for an application to certificate verification). This verification strategy is perhaps the most convenient, since it requires little infrastructure in addition to that used for training the neural certificate; however, generalization bounds tend to be conservative, and in some cases it may be desirable to provide firm (i.e. non-probabilistic) soundness guarantees.

The second common verification framework provides such deterministic guarantees and is based on the Lipschitz constant. Recall that a scalar function $f(x): \R^n \mapsto \R$ is said to be Lipschitz continuous with Lipschitz constant $L$ if for any two $x_1, x_2 \in \R$, it holds that $\abs{f(x_1) - f(x_2)} \leq L \norm{x_1 - x_2}$ for an appropriate norm. Certificate-learning works that make a Lipschitz argument for verification (e.g.~\cite{Richards2018,Bobiti2018}) compute the Lipschitz constant for each certificate constraint $L_{c_i}$, and then check the margin of constraint satisfaction at each point on either a fixed-size~\cite{Richards2018} or adaptive~\cite{Bobiti2018} grid. If the constraint is satisfied with some margin (i.e. $c_i(x_j, V) \leq -L_{c_i} \tau$, where $2\tau$ is the distance between adjacent grid points), then the constraint is guaranteed to be satisfied continuously between those points. This strategy is convenient, since it can be applied to any Lipschitz continuous constraint without much overhead. However, it has a number of drawbacks. First, the use of the Lipschitz constant necessarily introduces conservatism by assuming worst-case variation between sampled points (and the estimated Lipschitz constant itself can be either difficult to estimate~\cite{fazlyab_lipschitz_sdp} or provide a very loose upper bound). Second, checking the constraints over a grid of points incurs the curse of dimensionality, limiting applications to higher-dimensional systems. Third, this after-the-fact verification scheme does little to guide the training process, since it is only used after training is complete (this drawback applies to both Lipschitz arguments and generalization error bounds).

\begin{figure}[h]
    \centering
    \includegraphics[width=0.5\linewidth]{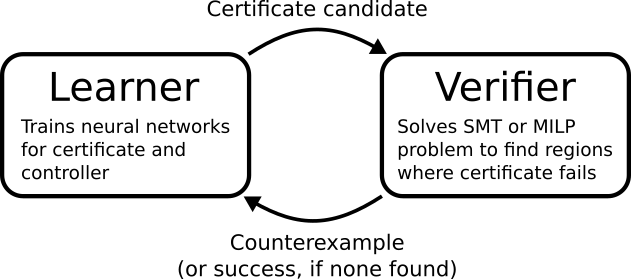}
    \caption{A learner-verifier architecture. Several different technologies may be used for the verifier, including SMT and MILP solvers.}
    \label{fig:learner_verifier}
\end{figure}

The third common category of verification methods, based on optimization, fills this gap by running periodically throughout the training process to verify the certificate as it is learned (and provide feedback in the form of counterexamples where the certificate is not yet valid). These approaches typically take the form of a \textit{learner-verifier} architecture, sometimes also referred to as counterexample-guided inductive synthesis (CEGIS~\cite{Chang2019,Abate2020,Peruffo2020,Dai2020,abate_fossil}). In these architectures (illustrated in Fig.~\ref{fig:learner_verifier}), the learner is responsible for training the certificate neural network, while the verifier periodically checks the learned certificate. If the certificate is valid, the verifier stops the training early; otherwise, it provides counterexamples to enrich the training dataset.

Depending on how the certificate and system dynamics are formulated, the verifier can be implemented using a number of different technologies. If the underlying dynamics are piecewise affine and all neural networks are encoded using Rectified Linear Unit (ReLU) activation functions, then the verification problem can be represented as a mixed-integer linear program (MILP) and solved using off-the-shelf MILP solvers, as proposed in \cite{Dai2020}. If the underlying dynamics are nonlinear, then the verification problem can be solved using satisfiability modulo theory. SMT problems can be solved using dedicated solvers such as dReal \cite{Gao2013}. For examples of this approach, see \cite{Chang2019,Abate2020,Peruffo2020,abate_fossil,Ahmed2020}.

Both MILP- and SMT-based methods help address the certificate verification problem, with the added benefit of speeding the training process by providing counter-examples. However, both methods are computationally expensive, and their complexity typically grows exponentially with the number of neurons in the certificate network. In practice, this poor scalability limits the applicability of these methods; for example, the MILP-based method in \cite{Dai2020} is demonstrated on networks containing only 16 neurons. More advanced MILP-based neural network verification tools exist, such as MIPVerify \cite{Tjeng2017}, but empirical results suggest they are currently limited to networks with less than 200 neurons \cite{Liu2019}. SMT-based methods have been demonstrated on networks of up to 30 neurons in \cite{Peruffo2020}. It is possible that recent advances in neural network verification (see \cite{Liu2019} for a relevant survey) or adversarial analysis and training of neural networks~\cite{madry2018towards} may improve the scalability of verifier architectures; however, most approaches are currently limited to small neural networks.

As a final note on the topic of verification, it is important to note that control certificates can still be valuable without exhaustive verification, for two reasons. First, even non-verified certificates can act as useful supervision for training control policies~\cite{Chang2021}. Second, it is possible to deploy a non-verified Lyapunov or barrier certificate safely using a real-time safety monitor. This monitor tracks the rate of change of the certificate and checks whether the derivative condition~\eqref{lyap_decrease_semi} (for Lyapunov functions) or~\eqref{bf_decrease} (for barrier functions) is violated; any such violation indicates that the certificate is no longer able to guarantee stability or safety and the system should enter a fail-safe mode.

\subsection{Effects of actuator limits}
Similarly to constrained model-predictive control, certificate-based controllers such as the quadratic program given in~\eqref{clf_qp_controller} can easily adapt to actuator limits \cite{ames_cbf}. The control policy is simply modified to respect the actuator limits (a CLF-based policy is shown here for example):
\begin{align}
    \min_{u \in \mathcal{U}} &\ ||u||^2 \label{clf_qp_controller_u_lim}\\
    \text{s.t.} &\ L_f V(x) + L_g V(x) u \leq -cV(x) \\
    &\ u \in \mathcal{U}
\end{align}
where $\mathcal{U}$ is the set of admissible controls. When $\mathcal{U}$ is a polytope, then this policy remains a quadratic program and can be solved efficiently online. The only complication is ensuring that this QP will be feasible. A neural certificate architecture can accommodate this constraint through the use of differentiable convex programming \cite{cvxpylayers2019}, which allows back-propagation through a the result of solving a quadratic program like~\eqref{clf_qp_controller_u_lim}. To train the certificate network in this context, the QP policy is relaxed:
\begin{align}
    \min_{u,r} &\ ||u||^2 + r^2 \label{clf_qp_controller_u_lim_relaxed}\\
    \text{s.t.} &\ L_f V(x) + L_g V(x) u \leq -cV(x) + r \\
    &\ u \in \mathcal{U} \\
    &\ r \geq 0
\end{align}
and the certificate network is trained with an additional loss term to minimize $r$; see~\cite{dawson2021safe} for an example. Although solving this optimization problem at training-time incurs an additional offline cost, it enables a learning-enabled controller to provide assurance that it will respect actuation limits at runtime (as contrasted with black-box learned policies such as those derived from reinforcement learning, which make no such guarantees).

\subsection{Attainable control frequencies}

Another important consideration when deploying controllers to hardware is the computational burden of running that controller in real-time. This can become an issue when the complexity of the control task increases; for example, it can be difficult to run robust model-predictive control (MPC) algorithms in real-time, due to the additional complexity of considering the effect of disturbances over a long horizon \cite{dawson2021safe}. In the case of MPC in particular, the need to consider the vehicle's safety over a multi-step horizon imposes a significant computational cost. In contrast, certificate-based controllers effectively encode long-term system properties (like safety and stability) into local properties of the certificate function; synthesizing a certificate can be seen as \textit{compiling} long-term behaviors into local properties. As a result, certificate-based controllers need consider only a single-step horizon when evaluating a controller such as~\eqref{clf_qp_controller}. Empirically, this results in at least an order of magnitude increase in attainable control frequency as compared to robust MPC~\cite{dawson2021safe}. Because a neural certificate control framework shifts the burden of computation to the offline stage (synthesizing the certificate), it is well suited for deployment on robots with limited computational resources or in situations where high control frequencies are required.

\section{Case Studies}\label{case-studies}

In this section, we present a series of instructive examples to demonstrate how the certificate-learning techniques discussed in this survey can be applied to practical problems in robotics. The first example demonstrates a straightforward application of certificate learning: given a nonlinear model of an autonomous car with uncertainty, we synthesize a trajectory-tracking controller that is robust to variation in the reference trajectory. Secondly, we demonstrate combining a learned CLF with a learned CBF to synthesize an observation-feedback controller for a mobile robot navigating a cluttered, unknown environment. We also include results from a hardware demonstration for this second example.

Source code for each of these examples, along with pre-trained models, are available on the accompanying website: \url{https://github.com/MIT-REALM/neural_clbf}. This code is intended to serve as a well-documented reference implementation for many of these techniques, and we hope that they will provide a useful starting point for any readers interested in applying these techniques to their own problems.

\subsection{Learning a CLF for complex dynamics}

Our first example involves learning a stabilizing tracking controller for a nonlinear model of an autonomous car. We use the single-track car model from the CommonRoad benchmarks~\cite{Althoff2017a}. This model includes not only nonlinear steering dynamics, but also the effects of friction and load transfer between the front and rear axles as the car accelerates. The model includes 7 state dimensions and two control inputs: steering angle velocity and longitudinal acceleration. More details on the dynamics can be found in~\cite{Althoff2017a}, or as implemented in the code accompanying this paper. We modify the dynamics from~\cite{Althoff2017a} slightly to express the car's state relative to a reference trajectory. This example builds on experiments presented in~\cite{dawson2021safe}.

These dynamics are not amenable to traditional control synthesis techniques (such as SoS, which fails due to numerical issues even when approximating these dynamics using polynomials~\cite{dawson2021safe}). However, we can apply the techniques introduced in Section~\ref{learning} to learn a controller based on a Control Lyapunov Function (CLF) that stabilizes this system about the reference trajectory. This section will explain how we apply these techniques to this specific problem.

To simplify the learning problem, we look for a CLF, which defines the optimization-based control policy~\eqref{clf_qp_controller} and removes the need to simultaneously search for a control policy (in effect, the control policy is parameterized by the CLF). We then instantiate the optimization problem~\eqref{eq:generic_certificate_search} with the specific constraints governing CLFs, given in~\eqref{clf_condition_goal}--\eqref{clf_condition}. We encode the CLF with learnable parameters $\theta$ as $V_\theta(x) = w_\theta(x)^T w_\theta(x)$, where $w_\theta$ is a fully-connected neural network with $\tanh$ activation functions and size $9 \times 64 \times 64$ (with the car's 7 state dimensions expanded to 9 by replacing two angular dimensions with their sine and cosine). To improve the convergence of the learning process, we linearize the dynamics, construct a LQR controller for the linearized system, and find the Lyapunov function for the closed-loop linear system $V_{lin} = \frac{1}{2}x^T P x$, where $P$ is the solution to the Lyapunov matrix equation. This linear solution is used to initialize the learned CLF by training the learned CLF to imitate $V_{lin}$ for 11 epochs; once this initialization phase is complete, we do not reference the linearized Lyapunov function and train to satisfy the constraints~\eqref{clf_condition_goal}--\eqref{clf_condition} (training continues to 25 total epochs).

Our choice to express the vehicle's state relative to the reference trajectory introduces a new challenge to this leaning problem. We wish to ensure that our learned CLF is valid not only for the reference trajectories seen during training, but also for the range of trajectories we might expect to see in practice. To address this issue, we can draw on work creating robust variants of CLF certificates~\cite{dawson2021safe}: we select two ``scenarios'' characterizing the uncertainty in the reference trajectory (representing maximum steering effort to the right and left), then train the CLF to be valid in both of these scenarios. Using our knowledge of the underlying model and the structure of the CLF conditions~\eqref{clf_condition} (which are affine in the model uncertainty, as the authors of~\cite{dawson2021safe} demonstrate), we can prove that the trained controller generalizes from these scenarios to more general trajectories.

\begin{figure}[h]
    \centering
    \includegraphics[width=0.8\linewidth]{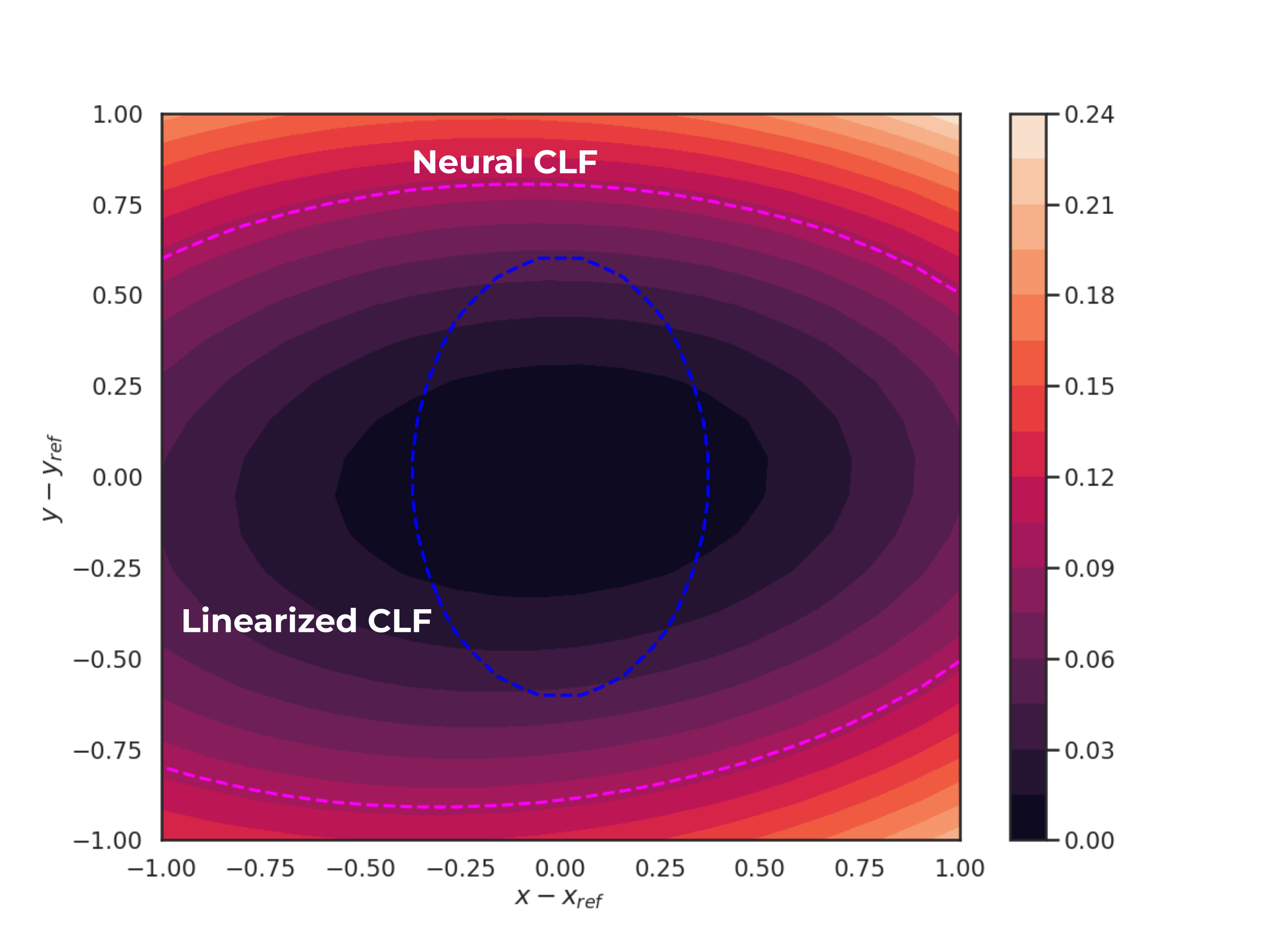}
    \caption{The CLF learned for the single-track car. Lighter colors indicate higher values of $V$; as expected, $V$ decreases to zero at the origin (corresponding to zero error relative to the reference trajectory). By ensuring that trajectories flow "downhill" on this landscape, a CLF-QP control policy can robustly track the reference trajectory. The $0.1$-level sets of the neural CLF and the CLF found via linearization are overlaid to highlight the difference.}
    \label{fig:stcar_clf}
\end{figure}

\begin{figure}[h]
    \centering
    \includegraphics[width=\linewidth]{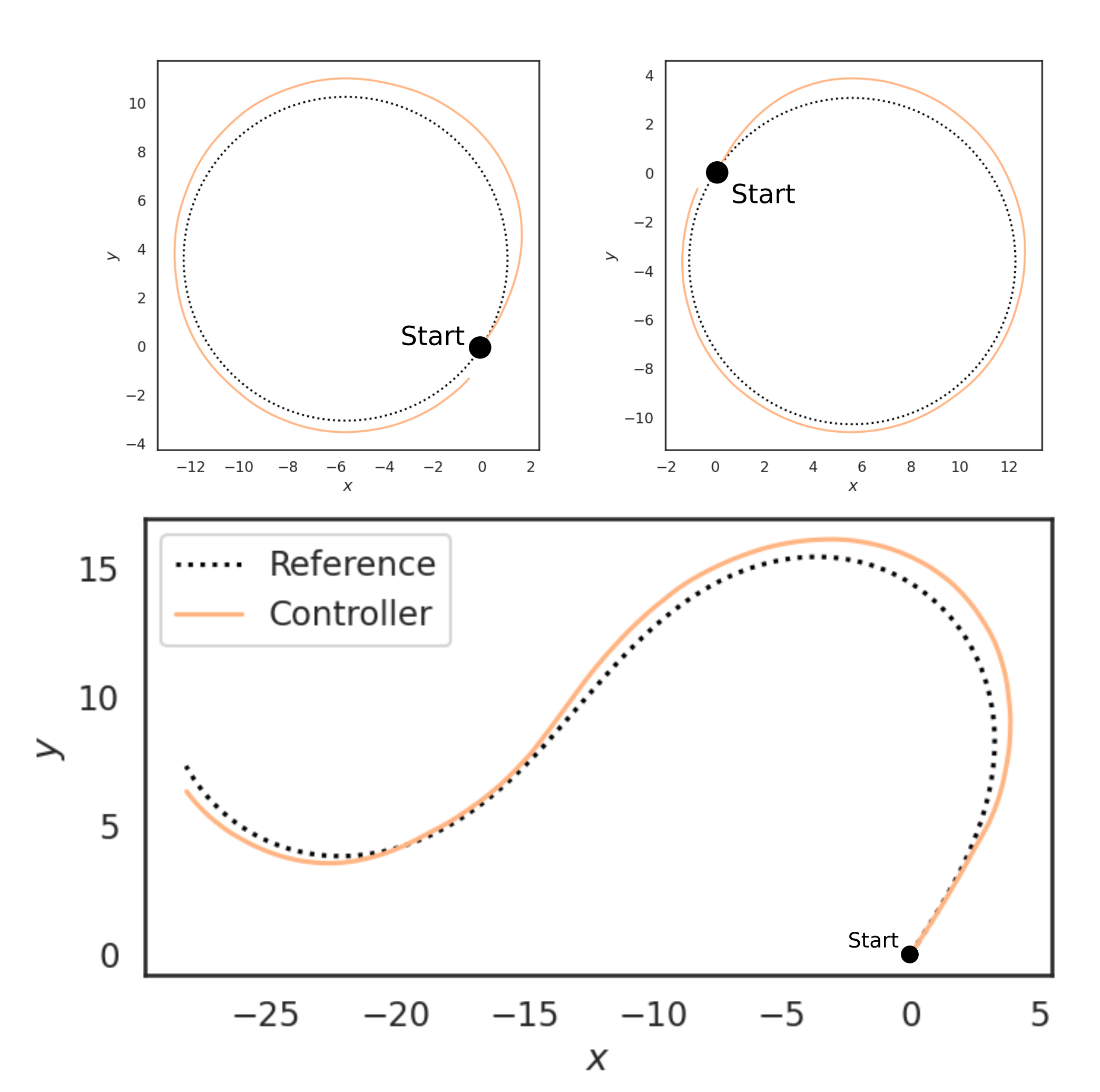}
    \caption{Tracking performance of a controller based on a learned CLF. Note that the system is only trained on ``scenarios'' like the circular reference paths shown above, but it successfully generalizes to more general reference paths through the use of the CLF.}
    \label{fig:stcar_tracking}
\end{figure}

A CLF for this system can be trained by running the code in \texttt{training/train\_single\_track\_car.py}, and a pre-trained model can be evaluated using the code in \texttt{evaluation/eval\_single\_track\_car.py}. The CLF learned using this method is shown in Figure~\ref{fig:stcar_clf} and the tracking performance is shown in Figure~\ref{fig:stcar_tracking}. Figure~\ref{fig:stcar_clf} clearly shows the difference between the neural CLF and the CLF found by continuous-time Lyapunov equation for the linearized system dynamics.


\subsection{Safe visual-feedback control with barrier functions}\label{case2}

Our second example demonstrates the application of certificates to ``full-stack'' robotics problems that integrate perception and control. In particular, we can combine a CBF with a CLF (both learned using neural networks) to certify the safety and liveness of a perception-feedback controller capable of safely navigating previously-unseen environments. At a high level, this system uses a CBF to ensure that it remains safe at all times, and it uses a CLF to guide itself to the goal. To avoid getting stuck when the CLF and CBF conflict, the controller is capable of temporarily suspending the CLF as needed to move around obstacles (this corresponds to tracing a level set of the CBF, i.e. maintaining a constant level of safety, while seeking a state from which the CLF can safely decrease). To enable generalization to new environments, the CBF and CLF are learned as functions of observations (Lidar data and range and bearing to the goal) rather than states. More details on this example can be found in~\cite{dawson22perception}. This example shows how learned certificates can be used to ensure safety as part of a larger robotics architecture.

We consider an autonomous ground robot with discrete-time nonlinear Dubins car dynamics and the ability to observe its environment via local Lidar observations, represented as range $o^i$ along each ray ($i=1,\ldots,N_r$). We also assume that the robot can measure its range $\rho$ and bearing $\phi$ relative to a beacon at the goal. We define the CBF $h(o) = h_\sigma(o) - \min_i ||o^i|| + d_c$, where $h_\sigma$ is a permutation-invariant neural network with parameters $\sigma$ and the second and third terms impose the prior that the barrier function should correlate with distance to the nearest obstacle. The CLF is represented as $V(\rho, \sin\phi, \cos\phi) = V_\omega(\rho, \sin\phi, \cos\phi) + \rho^2 + (1-\cos\phi)/2$, where $V_\omega$ is a neural network and the second and third terms impose the prior that the Lyapunov function should correlate with distance to the goal. These neural networks are trained in a single randomly-generated 2D environment using the self-supervision method discussed in Section~\ref{learning}.

Although the state dynamics of this system are control-affine, as discussed in Section~\ref{background}, the dynamics in observation space are not. As a result, we cannot use a simple quadratic-program control policy like Eq.~\eqref{clf_qp_controller}; instead, we must find a control input that satisfies the certificate conditions~\eqref{V_exp_decrease} and~\eqref{bf_decrease} by searching over the action space directly. This is computationally feasible, e.g. by discretizing the action space, but extending this approach to higher-dimensional action spaces, perhaps by applying stochastic optimization methods such as CMA-ES~\cite{kochenderfer_wheeler_2019}, is an interesting area of future work. Finally, we must handle the complication that arises when the system cannot simultaneously preserve safety by satisfying the barrier function condition~\eqref{bf_decrease} while moving towards the goal (satisfying the Lyapunov condition~\eqref{V_exp_decrease}). To do this, we can construct a simple hybrid controller that switches between goal-seeking and exploratory modes, preserving safety in all modes by enforcing the barrier function conditions, but able to temporarily disable the Lyapunov conditions in order to avoid getting stuck in local minima. In addition, the controller presented in~\cite{dawson22perception} defines a constrained stochastic control policy in the exploratory mode to encourage exploration of the state space.
The details of this switching strategy and a proof of its liveness can be found in~\cite{dawson22perception}.

\begin{figure}[h]
    \centering
    \includegraphics[width=\linewidth]{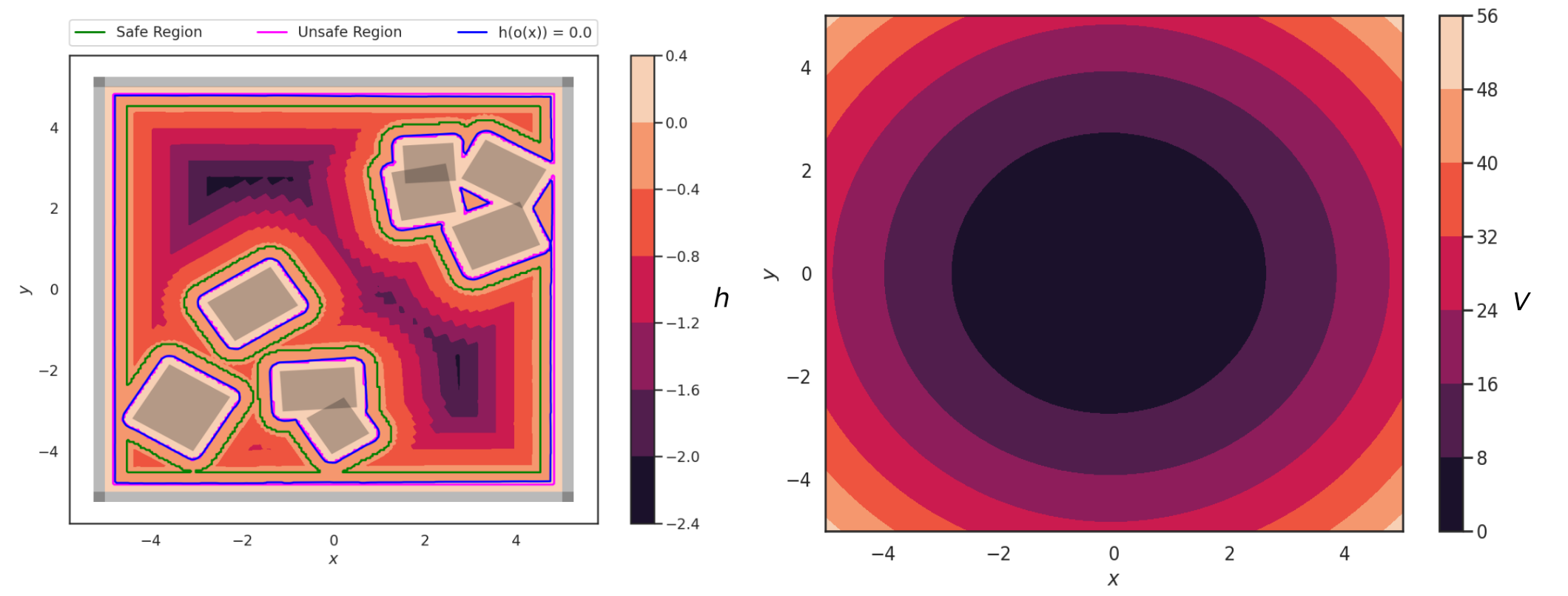}
    \caption{The CBF $h$ (left) and CLF $V$ (right) learned for navigating a cluttered environment (lighter colors indicate higher values). The learned CBF correlates with distance to the nearest obstacle, while the learned CLF correlates with distance from the goal. Since the CBF is learned as a function of observations, it can easily generalize to new environments.}
    \label{fig:cbf_nav}
\end{figure}

\begin{figure}[h]
    \centering
    \includegraphics[width=\linewidth]{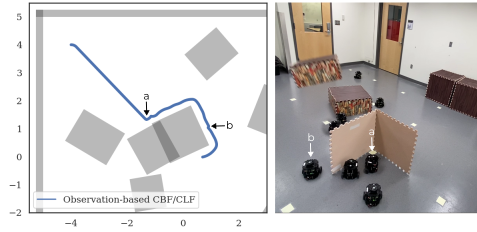}
    \caption{A certificate-based hybrid perception-feedback controller navigating a previously-unseen environment in both simulation (left) and hardware (right). In the hardware demonstration, a new obstacle is thrown in front of the robot partway through the experiment; the CBF allows the robot to gracefully avoid this obstacle. Labels (a) and (b) indicate where the policy switches from goal-seeking to exploratory and vice-versa, respectively.}
    \label{fig:cbf_nav_rollout}
\end{figure}

The learned CBF and CLF are shown in a random environment in Fig.~\ref{fig:cbf_nav}, and the hybrid controller is shown navigating a previously-unseen environment in Fig.~\ref{fig:cbf_nav_rollout}. Even though these certificates were trained using a single randomly-generated environment, the controller is able to navigate new environments safely without adapting its certificate. To demonstrate the potential of certificate-enabled controllers to handle some of the implementation challenges in Section~\ref{implementation}, we also include results deploying this controller on hardware in a laboratory environment in Fig.~\ref{fig:cbf_nav}.


\section{Limitations and Future Work}\label{future_work}

In our opinion, certificate-based learning for control has great promise but a number of drawbacks, which we discuss in this section. In addition to these limitations, we also highlight promising directions for future research in this area.

\subsection{Limitations}

\subsubsection{Data requirements \& Generalization}

Presently, the field's understanding of the amount of data required to successfully train a neural certificate is based largely on intuition and experience; there is not yet a firm theoretical footing to predict the data required by a certificate-learning framework. Some works have made some promising progress in this direction; notably, Boffi \textit{et al.}~\cite{Boffi2020} prove that certificate learning is \textit{asymptotically consistent} in the sense that as more data points are added to the training set, the maximum volume where the trained certificate fails to be valid shrinks to zero, but their analysis is restricted to systems that are known \textit{a priori} to be stable. It remains to extend this theory to the case of general nonlinear systems, or systems where the controller is learned alongside a certificate.

Closely related is the issue of generalization error, which relates a learned certificate's performance on a finite training set with its performance on the full state space (as discusses in Section~\ref{certificate_verification}). Some works~\cite{Boffi2020,Qin2021} have applied statistical learning theory to provide probabilistic upper-bounds on the generalization error (i.e. upper bounds that hold with some high probability), but these bounds tend to be conservative. There have been some preliminary attempts to apply almost-Lyapunov theory~\cite{Liu2020} to certificate learning~\cite{Chang2021}, but it remains to formalize this connection.

\subsubsection{Scalable verification}

As mentioned in Section~\ref{certificate_verification}, there has yet to be an effective solution to the certificate verification problem that scales to networks involving $> 100$ neurons. Probabilistic guarantees based on generalization error bounds~\cite{Boffi2020,Sun2020,Qin2021} or almost Lyapunov theory~\cite{Liu2020} have the potential to scale by replacing exhaustive verification with probabilistic guarantees, but this theory has yet to be fully developed and integrated into the certificate learning process (all existing applications of these techniques are \textit{post hoc}).

\subsection{Future work}

In addition to addressing these limitations in the existing theory of certificate learning, we believe that there are several promising directions to extend this theory to more complex controls and robotics problems.

\subsubsection{Model-free certificate learning and theoretical connections to RL}

Although works like~\cite{qin2021_sablas} and~\cite{Chang2021} deploy certificate learning in the black-box dynamics model and unknown-model settings, respectively, there remains much room to explore both practical and theoretical connections between certificate learning and reinforcement learning. On a practical level, \cite{Chang2021,westenbroek_2020,WESTENBROEK202119} all propose to learn the parameters of a Lyapunov function~\cite{Chang2021}, CLF~\cite{westenbroek_2020}, or CLF and CBF together~\cite{WESTENBROEK202119} via reinforcement learning.

On a theoretical level, deep connections between certificate learning and reinforcement learning remain unexplored. For example, \cite{berkenkamp17saferl} shows that it is possible to construct a reward function for specific RL problems so that the corresponding value function is also a Lyapunov function, but it is not well understood whether this is possible more generally. In particular, what conditions must the reward function satisfy so that the value function implies a Lyapunov function? Is it always possible to construct such a reward, even when non-stability objectives are included in the reward function? Work on constrained Markov decision processes (cMDPs) suggests that it is possible to derive Lyapunov functions from a cMDP's cost metric~\cite{chow2018lyap_rl}, and later work~\cite{chow2019lyapunovbased} has extended this approach to continuous state and action spaces. We anticipate that these questions will be the basis for exciting future research at the intersection of RL and certificate learning.

\subsubsection{Heterogeneous multi-agent certificates}

With an eye towards deploying large fleets of autonomous robots (e.g. in drone delivery or driving contexts), some works have applied certificate learning to systems with multiple agents~\cite{Qin2021,meng21_pedestrian}. However, these works focus exclusively on the case when every agent in the fleet has identical dynamics and constraints, as this allows a single certificate function to be learned for all agents at once. To extend these methods to heterogeneous multi-agent systems (e.g. a fleet of multiple autonomous delivery trucks, each with its own fleet of UAVs), it may be possible to apply compositional verification methods~\cite{Fan2017,ivanov21compositional,shen18compositional} to learn a certificate for various subsets of agents and combine them with rigorous safety guarantees.

\subsubsection{Distributed \& network control}

Related to the subject of heterogeneous multi-agent control, we believe there is an opportunity for fruitful research in learning certificates for distributed systems and networks. In addition to compositional verification techniques discussed above, this setting raises a number of interesting issues. The first issue is control and communication delay~\cite{wang14distributed}, which has been studied in the controls literature for standard control barrier functions~\cite{molnar2021safetycritical} but has not yet been applied in a certificate learning context. The second issue is fault tolerance~\cite{matni14resilience}, since a distributed control system should ideally be resilient to failures in individual nodes; this will likely require building on existing work in resilient control (e.g.~\cite{Bouvier2021}) to develop frameworks for resilient certificate learning. The third issue is scalability; contraction metrics in particular will likely require some decomposition techniques to avoid computing the eigenvalues of large matrices at training time. The application of certificate learning in this area also presents an interesting opportunity to apply graph neural networks~\cite{wu21gnn,yang2021communication} to learn network control certificates.

\section{Conclusion}\label{conclusion}

In this survey, we have reviewed an emerging suite of methods for automatically synthesizing safe controllers for nonlinear systems --- the neural certificate framework. This framework builds on established control theoretic concepts (Section~\ref{background}) by applying the representational power of neural networks (Section~\ref{learning}) to synthesize control certificates such as Lyapunov functions (Section~\ref{ex_lyap}), barrier functions (Section~\ref{ex_cbf}), and contraction metrics (Section~\ref{ex_cm}). We have also discussed several practical and theoretical challenges within this framework, as well as proposed tools from the literature to mitigate these challenges (Section~\ref{implementation}). Finally, we have presented two case studies (Section~\ref{case-studies}) and discussed directions for future work (Section~\ref{future_work}). We hope that this review provides an accessible jumping-off-point and high-level perspective on this emerging field.

A general framework for automatic certificate synthesis has long eluded control theorists. Since the development of Lyapunov functions in the late 1800s and modern control theory in the mid-to-late 1900s, a number of techniques have brought us progressively closer to this goal. LP-based simulation-guided synthesis~\cite{Kapinski2014} and sums-of-squares programming \cite{Ahmadi2016} provide important steps in this direction, each addressing some of the restrictions that applied to earlier works. In this context, neural certificates represent an important step forward for the state of the art for computational synthesis of nonlinear controllers. Neural certificates do not require a choice of hand-designed basis, as LP-based methods do, nor are they limited to systems with polynomial dynamics, as SoS methods are. Due to their generality, control architectures based on neural certificates have great promise for a wide range of robotics problems. Due to their generality and ease of implementation, we anticipate that neural certificates will see increasing adoption among practicing roboticists solving nonlinear safe control problems.

\bibliographystyle{IEEEtran}
\bibliography{IEEEabrv,main}

\FloatBarrier
\begin{table*}[htbp!]
\caption{Qualitative assessment of existing methods for controller safety certification.} \label{methods_comparison}
\begin{tabular}{ r c c c c }
    & Sum-of-Squares~\cite{Ahmadi2016} & CEGIS~\cite{Kapinski2014} & HJ Reachability~\cite{bansal_hji} & Neural certificates \\
    \hline
    Scalability & \parbox[t]{3.5cm}{$O(n^{6.5 d})$ for $d$-degree monomials~\cite{zhang_sdp_solver}} & SMT is NP-complete~\cite{Gao2013} & $O(\epsilon^{-n})$, grid size $\epsilon$~\cite{bansal_hji} & Requires sampling state space \\
    Robustness to disturbance & \cmark & \xmark & \cmark & \cmark \\
    Parametric uncertainty & \cmark & \xmark & \xmark & \cmark \\
    Model & Known polynomial & Known & Known & May be unknown~\cite{Castaneda2020,qin2021_sablas} \\ &&&&
\end{tabular}
\end{table*}



\begin{IEEEbiography}[{\includegraphics[width=1in,height=1.1in,clip,keepaspectratio]{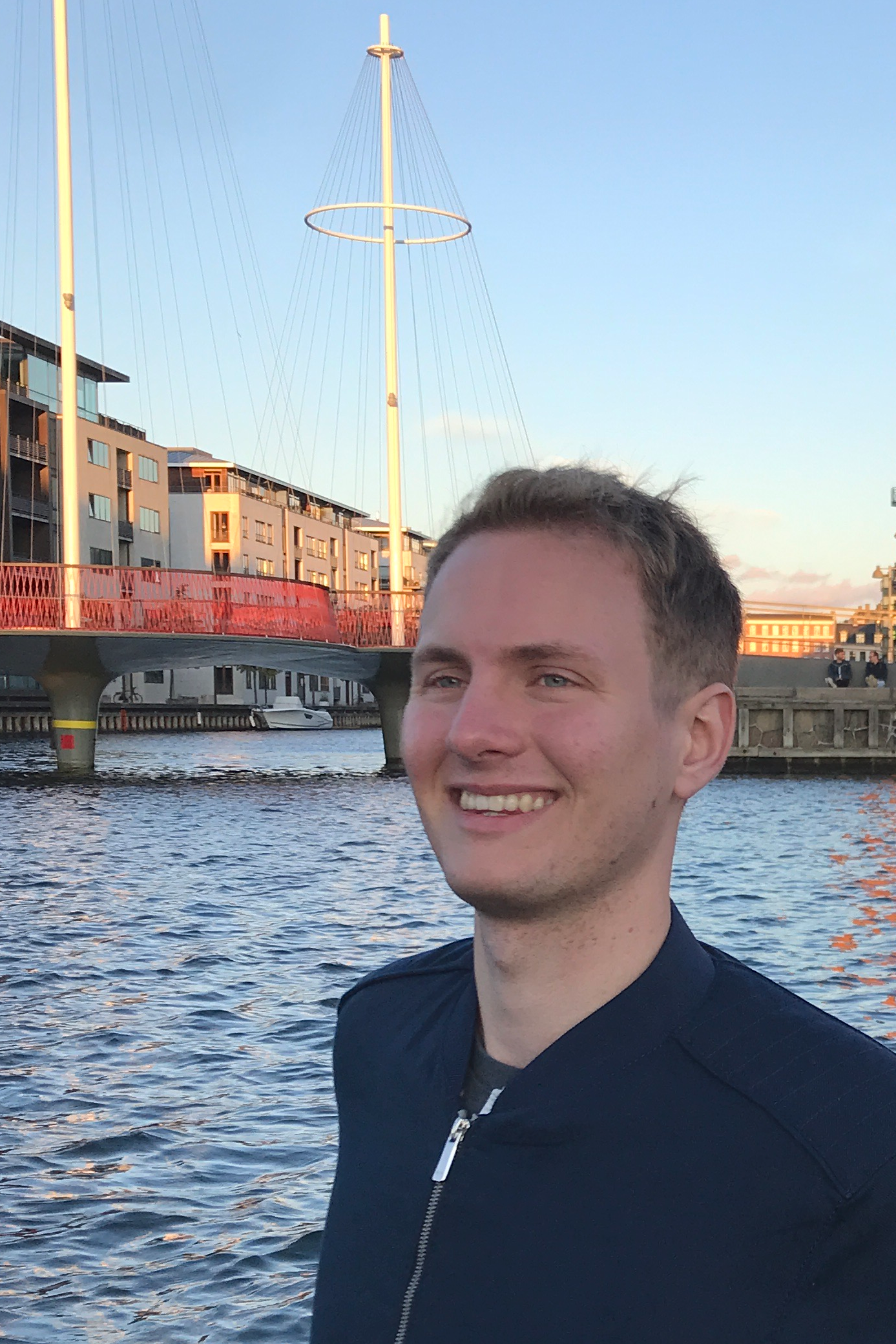}}]{Charles Dawson}
is a graduate student in the Department of Aeronautics and Astronautics at MIT, supported by the NSF Graduate Research Fellowship. He works on using tools from controls, learning, and optimization to understand safety in autonomous and cyberphysical systems. Prior to MIT, he received a B.S. in Engineering from Harvey Mudd College.
\end{IEEEbiography}

\begin{IEEEbiography}[{\includegraphics[width=1in,height=1.1in,clip,keepaspectratio]{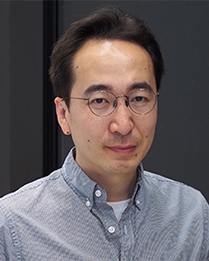}}]{Sicun Gao} is an Assistant Professor in Computer Science and Engineering at the University of California, San Diego. He works on computational methods and tools for improving automation and autonomous systems. He is a recipient of the Air Force Young Investigator Award, the NSF Career Award, and a Silver Medal for the Kurt Godel Research Prize. He received his Ph.D. from Carnegie Mellon University and was a postdoctoral researcher at CMU and MIT.
\end{IEEEbiography}

\begin{IEEEbiography}[{\includegraphics[width=1in,height=1.1in,clip,keepaspectratio]{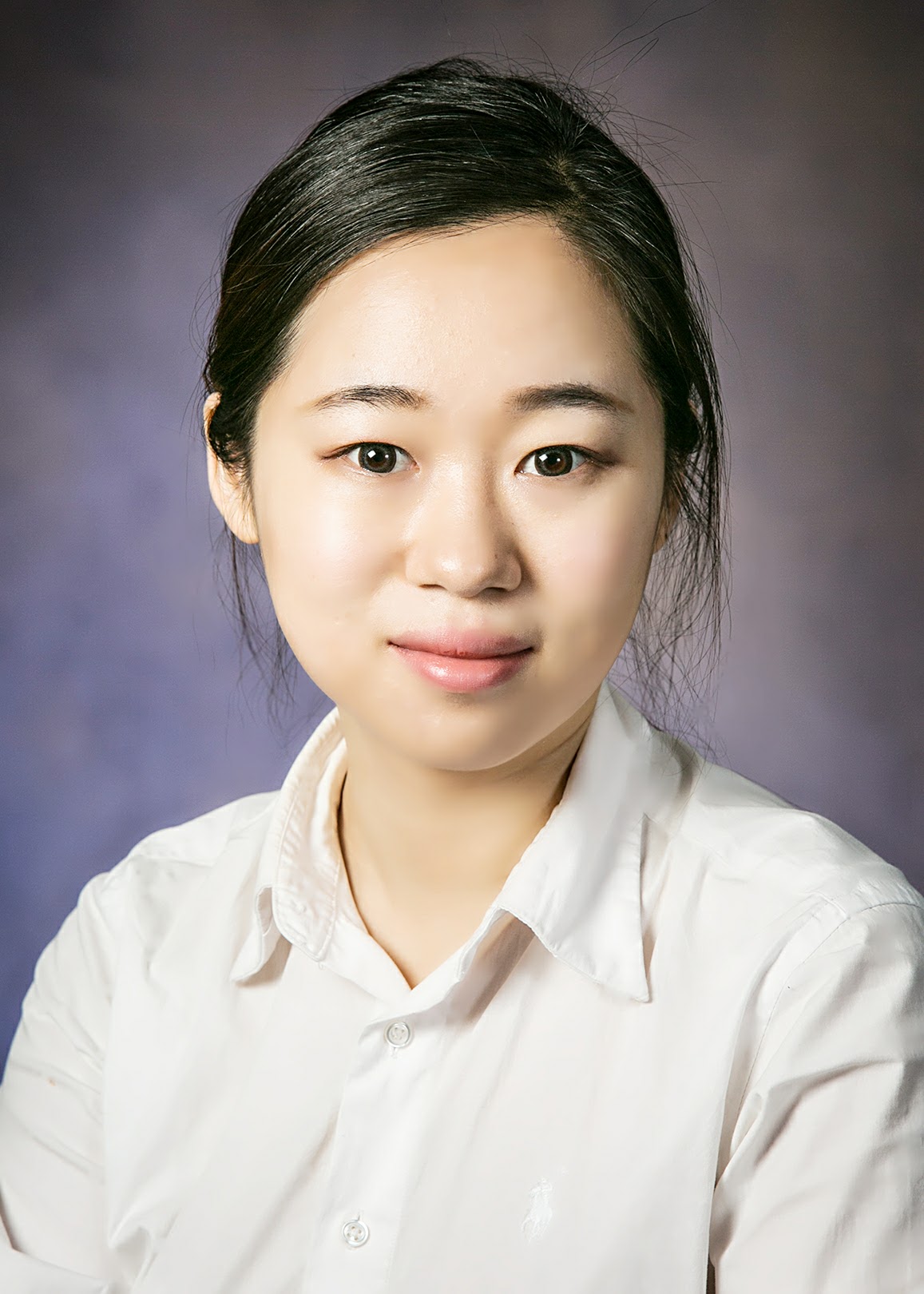}}]{Dr. Chuchu Fan} is an Assistant Professor in AeroAstro and LIDS at MIT. Before that, she was a postdoctoral researcher at Caltech and got her Ph.D. from the Electrical and Computer Engineering Department at the University of Illinois at Urbana-Champaign. Her group at MIT, REALM, works on using rigorous mathematics, including formal methods, machine learning, and control theory, to design, analyze, and verify safe autonomous systems. She is the winner of the AFSOR YIP Award, Innovator under 35 by MIT Technology Review, and the ACM Doctoral Dissertation Award.
\end{IEEEbiography}

\end{document}